\documentclass[lettersize,journal]{IEEEtran}
\usepackage{amsmath,amsfonts}
\usepackage{array}
\usepackage{textcomp}
\usepackage{stfloats}
\usepackage{url}
\usepackage{verbatim}
\usepackage{graphicx}
\usepackage{cite}
\usepackage{graphicx}
\usepackage{stfloats}
\usepackage{placeins}
\usepackage{subfigure}
\usepackage{amsmath}
\usepackage{mathtools}
\usepackage{float}
\usepackage{amsfonts,amssymb}
\usepackage{multirow}
\usepackage{makecell}
\usepackage{tabulary}
\usepackage{hyperref}
\usepackage{subcaption} % 必须的宏包，用于创建子图
\usepackage{booktabs} % For \toprule, \midrule, \bottomrule
\usepackage{siunitx}    % For decimal alignment and consistent unit formatting
\usepackage{threeparttable}
\usepackage{cuted}
\usepackage{makecell}
\usepackage{algorithm}
\usepackage{algpseudocode}
\usepackage{diagbox}
\hypersetup{hidelinks,
	colorlinks=true,
	allcolors=blue,
	pdfstartview=Fit,
	breaklinks=true}
\DeclareMathOperator*{\argmax}{arg\,max}
\begin{document}

\title{MAVEN: A Meta-Reinforcement Learning Framework for Varying-Dynamics Expertise in Agile Quadrotor Maneuvers}

\author{Jin Zhou$^{1}$, Dongcheng Cao$^{1}$, Xian Wang$^{1}$, and Shuo Li$^{1}$
\thanks{$^{1}$Authors are with the College of Control Science and Engineering, Zhejiang University, Hangzhou 310027, China
        {\tt\small j.zhou2020@zju.edu.cn}
        }
}

\maketitle

\begin{abstract}
Reinforcement learning (RL) has emerged as a powerful paradigm for achieving online agile navigation with quadrotors. Despite this success, policies trained via standard RL typically fail to generalize across significant dynamic variations, exhibiting a critical lack of adaptability. This work introduces MAVEN, a meta-RL framework that enables a single policy to achieve robust end-to-end navigation across a wide range of quadrotor dynamics. Our approach features a novel predictive context encoder, which learns to infer a latent representation of the system dynamics from interaction history. We demonstrate our method in agile waypoint traversal tasks under two challenging scenarios: large variations in quadrotor mass and severe single-rotor thrust loss. We leverage a GPU-vectorized simulator to distribute tasks across thousands of parallel environments, overcoming the long training times of meta-RL to converge in less than an hour. Through extensive experiments in both simulation and the real world, we validate that MAVEN achieves superior adaptation and agility. The policy successfully executes zero-shot sim-to-real transfer, demonstrating robust online adaptation by performing high-speed maneuvers despite mass variations of up to 66.7\% and single-rotor thrust losses as severe as 70\%. A supplementary video is available at [\href{https://youtube.com/playlist?list=PLbEQeDMEVpqzVj3aq1Otw4zweZKLcwtI7&si=weUw5HhTXBxG7mY2}{\textbf{\emph{video}}: https://youtube.com/playlist?list=PLbEQeDMEVpqzVj3aq1Otw-4zweZKLcwtI7}].
\end{abstract}

\section{Introduction}
\IEEEPARstart{T}{he} pursuit of autonomous agile flight for quadrotors has seen remarkable progress in recent years. While traditional optimization-based methods have laid a foundational groundwork \cite{foehn2021time}, \cite{wang2022geometrically}, recent breakthroughs have been increasingly driven by deep reinforcement learning (RL) approaches \cite{song2021autonomous}, \cite{end2024robin}. RL policies learn through extensive trial-and-error interactions in simulation. This paradigm allows agents to discover complex control strategies and execute aggressive maneuvers at the very edge of the quadrotor's physical limits, often rivaling or even surpassing the performance of expert human pilots \cite{kaufmann2023champion,song2023reaching}.

\begin{figure}
     \centering
    \includegraphics[width=0.5\textwidth,trim = 0 70 405 0, clip]{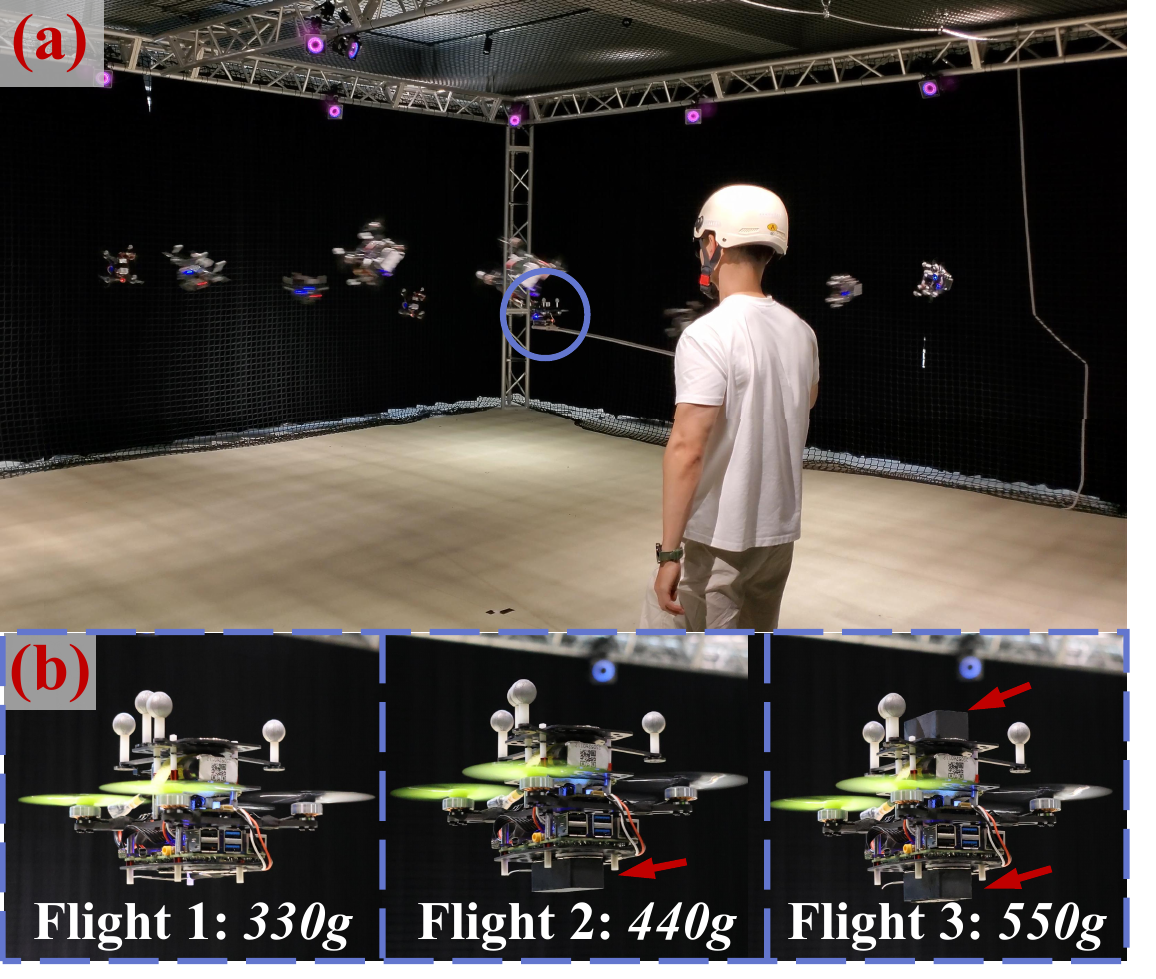}
    % \vspace{-1.0em}
    \caption{Demonstration of our policy performing three consecutive flights in the mass variation scenario without landing. (a) We alter the quadrotor mass with a magnet payload. (b) Illustration of a quadrotor with varying masses.}
    \label{real_mass_var}
\end{figure}

Particularly for agile navigation tasks based solely on sparse waypoints or high-level instructions, RL approaches represent a promising frontier. Implementing an end-to-end solution for such tasks is inherently beneficial, as it unifies the planning and execution phases to ensure rapid response and dynamic feasibility.  In realizing this integrated autonomy, RL approaches possess a critical advantage over traditional methods: computational efficiency. Although traditional optimization-based methods can theoretically generate trajectories that include feedforward control inputs, the intensive computation required often restricts them to offline generation \cite{foehn2021time}, \cite{qin2024time}. Consequently, they inevitably require a supplementary controller to compensate for tracking errors and dynamic environmental responses in real time \cite{sun2022comparative}, \cite{romero2022model}. In contrast, RL policies shift this heavy computational burden entirely to the offline training phase. During online deployment, the learned policy operates as a highly efficient neural network, enabling millisecond-level inference latency. This efficiency facilitates onboard execution on resource-limited micro quadrotors. Notably, recent studies have demonstrated this capability, achieving near-time optimal agile flight across multiple drones \cite{wang2025dashing} and successfully navigating unstructured forests using only an ARM-based computer \cite{zhang2025learning}.

In practice, a standard RL policy, trained for a specific set of system dynamics, excels in its nominal environment but often struggles to generalize to significant dynamic variations, such as drastic changes in vehicle configuration or unforeseen actuator faults. To overcome these limitations, domain randomization (DR) \cite{tremblay2018training}, \cite{molchanov2019sim} and fault-tolerant control (FTC) \cite{fei2020learn} emerge as effective methods to handle these variations. For instance, a recent study validates the effectiveness of the DR method in handling extreme dynamic variations in drone-racing scenarios, with a single policy capable of operating across different drones, even those with distinct sizes and propellers \cite{ferede2025one}. Recent studies also integrate DR with adaptive control or imitation learning, yielding low-level controllers that are robust to dynamic variations \cite{zhang2025learning,zhang2022learning}. Additionally, RL-based FTC extends standard RL methods, demonstrating remarkable success in handling motor failures \cite{fei2020learn}, \cite{kim2025reinforcement}. Notably, Liu et al. \cite{liu2024reinforcement} propose a hybrid hierarchical framework where an RL-based auxiliary controller augments a fundamental base controller. By learning to output compensatory signals that bolster the base controller, this approach effectively stabilizes the drone under actuator faults without requiring explicit fault modeling.

Despite their capabilities, these powerful methods have inherent limitations. While DR achieves broad robustness, it inevitably forces the policy to compromise task-specific optimality, resulting in a conservative policy with limited agility \cite{ferede2025one}. As a result, such a generalized policy exhibits varying degrees of performance degradation across different platforms, consistently falling short of the peak performance achieved by expert policies specialized for specific dynamics. Conversely, while FTC approaches demonstrate strong robustness against specific, anticipated faults, they are typically tailored to predefined failure models, thereby limiting its ability to generalize to unseen faults. More critically, current research in this domain focuses largely on the controller level, with validation restricted to position control or tracking fixed reference trajectories \cite{fei2020learn,kim2025reinforcement,liu2024reinforcement}. Consequently, there is limited exploration into fault tolerance at the trajectory planning level, where the flight trajectory itself must be re-optimized to accommodate the altered dynamics. These fundamental trade-offs between robustness and performance, and between specialization and adaptability, motivate our adoption of a paradigm that can learn to adapt.

To this end, we turn to extend the standard RL framework with meta-learning \cite{thrun1998learning,finn2017model,hospedales2021meta}, an approach known as meta-reinforcement learning (meta-RL) \cite{duan2016rl,rakelly2019efficient,bhatia2023rl}. The core idea is to train a policy from data across different tasks, enabling it to generalize and adapt rapidly to new, unseen ones. Meta-learning has proven highly effective across diverse robotics domains, ranging from legged robots with immediate adaptation on terrains and slopes \cite{nagabandi2018learning}, to wheeled robots under unknown operating conditions \cite{mckinnon2021meta}. A recent study further applies meta-RL to address actuator faults, enabling legged robots to master locomotion with motor stuck \cite{chen2024meta}.

The application of meta-learning to the quadrotor domain has seen primary success, demonstrating its potential for rapid adaptation \cite{wei2025meta}. Early works successfully integrate meta-learning principles into non-RL adaptive control frameworks. For instance, Neural-Fly \cite{o2022neural} learns a domain-invariant representation to estimate aerodynamic residuals, enabling stable flight under widely varying and rapidly changing wind conditions. Most recently, RAPTOR \cite{eschmann2025raptor} advances this direction by distilling thousands of expert policies into a single recurrent foundation policy via meta-imitation learning, achieving zero-shot adaptation across heterogeneous platforms. Furthermore, recent research begins to explore the specific application of meta-reinforcement learning (meta-RL). One such study applies model-based meta-RL to the suspended payload transport task \cite{belkhale2021model}, where the drone learns to rapidly adapt its internal dynamics model online to account for the unpredictable effects of unknown payloads and cable lengths. Similarly, another study develops a meta-policy for autonomous landing on a mobile platform, enabling adaptation to previously unseen landing trajectories  \cite{cao2024autonomous}.

However, while these studies validate the paradigm's potential, their application has been largely confined to low-level control, tracking predefined references \cite{wei2025meta,o2022neural,eschmann2025raptor,eschmann2024learning} or narrow task scopes \cite{cao2024autonomous}. This leaves the challenge of autonomous agile navigation largely unaddressed, particularly at the trajectory planning level, where the drone must actively re-optimize the trajectory to accommodate large dynamic discrepancies, rather than merely stabilizing a fixed path. Achieving this is challenging because existing meta-RL approaches often rely on real-world data \cite{o2022neural,belkhale2021model} or expert demonstrations \cite{eschmann2025raptor} that are expensive to acquire. Moreover, they typically suffer from prohibitive training times \cite{rakelly2019efficient,cao2024autonomous}. These combined limitations hinder the large-scale learning required for the complex trajectory planning level.

% high-level
To solve this problem, this work introduces MAVEN, a novel meta-RL framework designed specifically for agile flight across varying quadrotor dynamics. In contrast to prior works, our framework leverages high-fidelity simulation for massively parallel sampling to jointly optimize the policy network and the adaptation mechanism. This efficient process enables the rapid training of a robust navigation policy capable of direct, zero-shot deployment to physical platforms. Crucially, once deployed, the policy can rapidly adapt online to significant dynamic discrepancies. We validate this capability in two challenging scenarios: vehicle configuration variations and actuator faults. Through these scenarios, our approach demonstrates a combination of performance and robustness superior to that of the baseline methods. Overall, in this work:
\begin{enumerate}
    \item We propose a hybrid meta-RL framework for agile waypoint traversal that enables online adaptation to significant and unknown dynamic variations. The framework integrates an off-policy predictive context encoder for highly sample-efficient task inference with a stable on-policy PPO agent for motion planning, leveraging the respective strengths of both paradigms.
    \item We demonstrate that our method achieves both high agility and robust online adaptation across challenging, representative scenarios. The learned policy effectively handles large dynamic variations, such as significant mass variations and single-rotor thrust loss, in real-time.
    \item We validate the framework's zero-shot sim-to-real transfer, demonstrating through extensive experiments that a single policy, trained entirely in simulation, directly handles unseen dynamic variations on the physical quadrotor.
\end{enumerate}

% \section{Related Works}

\section{Methodology}

This section details our establishment of a navigation policy capable of agile quadrotor flight with significant, unobserved variations in system dynamics. We first formulate the waypoint traversal task under these uncertainties as a Partially Observed Markov Decision Process (POMDP). To solve this, we adopt a hybrid meta-RL framework that enables online adaptation. The core of our approach is a novel predictive context encoder, which learns to infer a latent variable representing the system dynamics. Finally, we elaborate on the efficient training and deployment procedure.

\subsection{Quadrotor Dynamics}
The state of a quadrotor is defined as ${\mathbf{x}} = [\mathbf{p}, \mathbf{v}, \mathbf{R}, \boldsymbol{\omega}]$, which includes the position, velocity, attitude (represented by a rotation matrix), and angular velocity, respectively. The dynamic model is described by the following equations:
% \vspace{-0.35cm}
\begin{align*}
    \dot{\mathbf{p}} &= \mathbf{v} & \dot{\mathbf{R}} &= \mathbf{R} \hat{\boldsymbol{\omega}} \\
    \dot{\mathbf{v}} &= \frac{1}{m}\mathbf{R}\mathbf{u} + \mathbf{g} & \dot{\boldsymbol{\omega}} &= \mathbf{J}^{-1}(\boldsymbol{\tau} - \boldsymbol{\omega} \times \mathbf{J}\boldsymbol{\omega})
\end{align*}
where $\hat{\boldsymbol{\omega}}$ is the skew-symmetric matrix of $\boldsymbol{\omega}$. The constants $m$ and $J$ are the quadrotor mass and inertia matrix, respectively, and $\mathbf{g}$ is the gravitational acceleration vector.

The total thrust vector $\mathbf{u}$ and the torque vector $\boldsymbol{\tau}$ are produced by the four individual rotor thrusts, ${u_i},i \in \{1,2,3,4\}$, subject to the constraint $0 \leq u_i \leq u_{max}$. The mappings from rotor thrusts to the collective thrust and torques are:
\begin{align*}
    {{\mathbf{u}}} = \begin{bmatrix}
        0 \\  0 \\ \sum {{u}_i}
    \end{bmatrix},
    {{\boldsymbol{\tau}}} = \begin{bmatrix}
        {{l}}/ \sqrt{2}( {u_1} + {u_2} -{u_3} -{u_4}) \\
       {{l}}/ \sqrt{2}( -{u_1} + {u_2} +{u_3} -{u_4}) \\
       {{c}}_{\tau}( {u_1} - {u_2} +{u_3} -{u_4})
    \end{bmatrix}
\end{align*}
where $l$ is the arm length and ${{c}}_{\tau}$ is the drag coefficient.

\subsection{Problem Statement} \label{prob state}
In this work, the objective is to develop a policy that navigates a quadrotor through a sequence of predefined waypoints, a task that encompasses both motion planning and control. To validate the effectiveness and robustness of our trained policy, we evaluate its performance across two challenging categories of scenarios: vehicle configuration variations and actuator faults.

Specifically, in the scenario with vehicle configuration variations, we vary the quadrotor mass $m$ within a predefined range of $[m_{min},m_{max}]$, while other parameters, such as the arm length and maximum thrust of the rotors, are held constant. In the actuator fault scenario, we introduce a loss of maximum thrust on a single, randomly selected rotor. The magnitude of this thrust loss, represented by a percentage factor $\delta_T$, is varied within the range $[\delta_{min},\delta_{max}]$.

By treating uncertainties like configuration variations or actuator faults as unobserved state components, we formulate the task of adaptive agile waypoint traversal as a POMDP. The main challenge of this POMDP is the information gap between the agent's observations and the actual system dynamics. To bridge this gap, the policy must infer unobserved dynamic parameters from its interaction history, since these parameters are not directly measurable.

To achieve this, we employ a meta-learning technique to infer and adapt to the unobserved system dynamics. The core is a probabilistic latent context variable vector $\mathbf{z} $, representing the unobserved system properties. This variable is inferred from historical information, or context $\mathbf{c}$, via a trained encoder network $q_{\phi}(\mathbf{z}|\mathbf{c})$. By conditioning the policy on this inferred variable, the original POMDP is effectively converted into a tractable Markov Decision Process (MDP).

We define the resulting MDP as a tuple $\mathcal{M}=(\mathcal{S}, \mathcal{A}, \mathcal{P}, \mathcal{R}, \gamma)$, where $\mathcal{S}$ denotes the state space, $\mathcal{A}$ the action space, $\mathcal{P}$ the state transition probability, $\mathcal{R}$ the reward function, and $\gamma$ the discount factor. The objective is to find an optimal policy $\pi_{\theta}^{*}$ that maximizes the expected discounted return:
\begin{equation}
\pi_{\theta}^{*} = \argmax_{\pi} ~\mathbb{E}\left[ \sum_{t=0}^{\infty} \gamma^t r_t \right]
\end{equation}
where $\gamma \in [0,1)$ is the discount factor and $r_t$ is the immediate reward at timestep $t$. $\mathcal{S}$, $\mathcal{A}$, $\mathcal{P}$ and $\mathcal{R}$ for this MDP are defined as follows:

\textbf{Observation Space:} The observation space $\mathcal{O}$ comprises all physical information perceivable to the agent. A single observation $\mathbf{o} \in \mathcal{O}$ consists of the quadrotor's velocity $\mathbf{v}$, its attitude as a flattened rotation matrix $\operatorname{vec}{\left(\mathbf{R}\right)}$, and the relative position vectors to the next $W$ waypoints, $\Delta \mathbf{p_k} \in \mathbb{R}^3$ for $k \in \{1,...,W\}$.

\textbf{State Space:} The state space $\mathcal{S}$ is an augmented space constructed to satisfy the Markov property. Each state $\mathbf{s} \in \mathcal{S}$ is a combination of the current physical observation and the inferred latent variable $\mathbf{z}$, defined as $\mathbf{s} = [\mathbf{o}, \mathbf{z}]$.

\textbf{Action Space:} The policy network $\pi_{\theta}$ generates a normalized action vector $\mathbf{a} \in \mathcal{A}$, defined as $\mathbf{a} = [\tilde{T}_{\text{cmd}}, \tilde{\boldsymbol{\omega}}_{\text{cmd}}]$. These normalized outputs are then transformed into physical commands for the flight control unit through the following mappings:
\begin{equation}
T_{\text{cmd}} = \tilde{T}_{\text{cmd}}, \quad
\boldsymbol{\omega}_{\text{cmd}} = \tilde{\boldsymbol{\omega}}_{\text{cmd}} \boldsymbol{\omega}_{\text{max}}
\end{equation}
where $T_{\text{cmd}}$ is the throttle command and $\boldsymbol{\omega}_{\text{max}}$ is the maximum body rates.

\textbf{Reward Function:} The reward function is designed to encourage agile waypoint traversal. The total reward at timestep $t$ is a weighted sum $r_t = \lambda_1 r_{\text{prog}} + \lambda_2 r_{\text{smooth}} + \lambda_3 r_{\text{safe}}$, where the components are defined as follows:
\begin{itemize}
    \item $r_{\text{prog}} = \|\Delta\mathbf{p}_{1,t-1}\|^2 - \|\Delta\mathbf{p}_{1,t}\|^2$: To promote minimum-time flight, we incentivize progress towards the next waypoint, defined as the reduction in squared Euclidean distance. This dense reward signal is calculated at each step to encourages the agent to close the distance to the immediate waypoint.
    \item $r_{\text{smooth}} = -\|\mathbf{a}_{t-1} - \mathbf{a}_{t}\|$: To promote dynamically feasible trajectories, we penalize abrupt changes in consecutive actions. This encourages control smoothness and mitigates high-frequency oscillations in the policy's output, which is critical for stable execution on physical actuators.
    \item $r_{\text{safe}}$: To ensure safety and penalize crash states, a large constant penalty, $r_{\text{safe}}=-r_{\text{collision}}$, is applied when a collision is detected. A collision is registered if the quadrotor exits the predefined valid workspace, an event which immediately terminates the episode.
\end{itemize}

\begin{figure*}
     \centering
    \includegraphics[width=0.99\textwidth,trim = 0 170 0 50, clip]{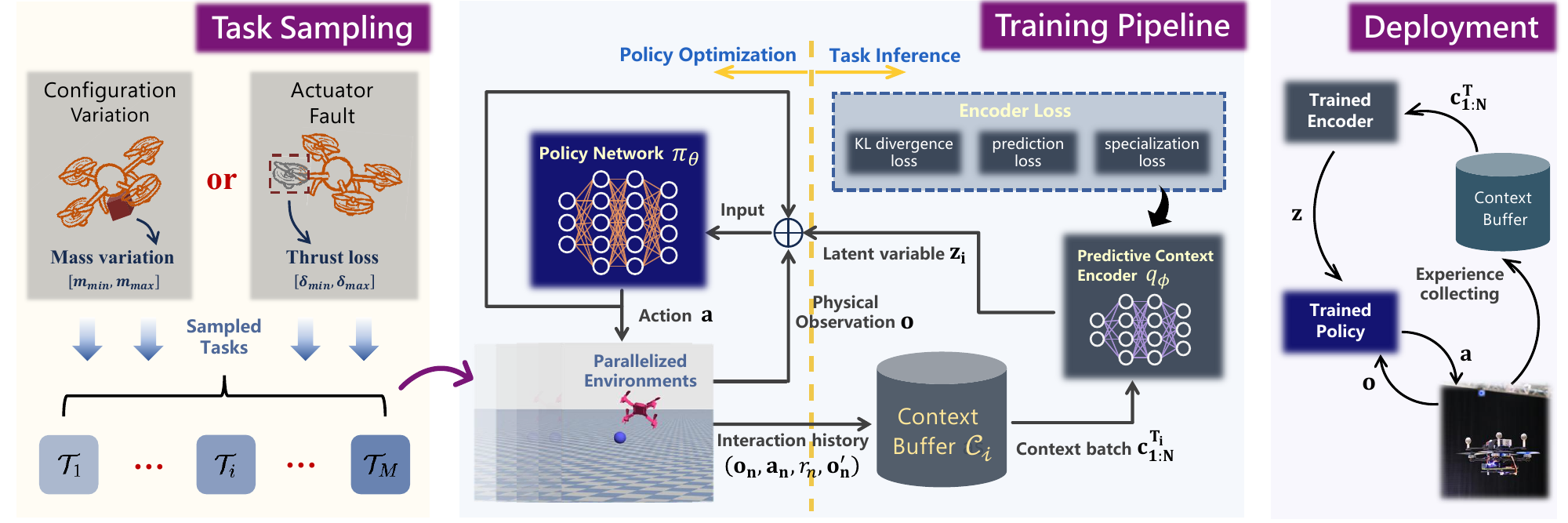}
    \caption{Overview of our meta-RL framework for online adaptation to quadrotor dynamics. The policy is trained in parallelized simulation environments on tasks with varying dynamics (mass and thrust loss). A novel predictive context encoder learns to infer a latent variable that conditions the policy network, enabling task-aware adaptation. For deployment, the trained policy performs real-time task inference and navigation on an onboard computer.}
    \label{training pipeline}
    % \vspace{-1.5em}
\end{figure*}

\subsection{Meta-RL Framework}\label{Meta-RL}
% Meta-RL + Inner loop
Inspired by PEARL \cite{rakelly2019efficient}, our meta-RL framework integrates an off-policy probabilistic context encoder with the on-policy Proximal Policy Optimization (PPO) algorithm \cite{PPO}, as shown in Fig. \ref{training pipeline}. While PEARL utilizes a fully off-policy framework, our framework instead adopts an on-policy agent for policy optimization. This integration is achieved by conditioning both the actor and critic networks on the inferred latent variable $\mathbf{z}$ in addition to the physical observation $\mathbf{o}$, thereby making the policy task-aware. This hybrid framework leverages the respective strengths of both paradigms: the high sample efficiency of off-policy learning for the task inference, and the stable policy updates of on-policy learning for the motion planning. Accordingly, our approach decouples the overall problem into two components: task inference and policy optimization, which are detailed below.

\subsubsection{Task inference}
First, we define a task $\mathcal{T}$ as a unique Markov Decision Process (MDP) corresponding to a specific set of system dynamics. Variations in the quadrotor mass or the actuator thrust loss alter these dynamics and their state transition probabilities, causing identical control commands to result in different flight behaviors for each task. By sampling from a distribution of these dynamic parameters, we generate a set of distinct tasks $\{\mathcal{T}_i\}$ for meta-training.

Task inference is performed by a probabilistic context encoder. To facilitate this, for each task $\mathcal{T}_i$, we maintain a separate off-policy context buffer $\mathcal{C}_i$, storing the interaction history as transition tuples $\mathbf{c}^{\mathcal{T}_i}_n=(\mathbf{o}_n,\mathbf{a}_n,r_n,\mathbf{o}'_n)$. During inference, we sample a batch of context, $\mathbf{c}^{\mathcal{T}_i}_{1:N} = \{\mathbf{c}_1^{\mathcal{T}_i}, \dots, \mathbf{c}_N^{\mathcal{T}_i}\}$, from its corresponding buffer $\mathcal{C}_i$. The encoder network, $q_{\phi}(\mathbf{z}|\mathbf{c}^{\mathcal{T}_i}_{1:N})$, then processes this batch to infer a posterior distribution over the latent variable $\mathbf{z}$. Since an MDP is fully characterized by an unordered set of its transitions, the inference process is designed to be permutation-invariant. Adopting the probabilistic structure from PEARL \cite{rakelly2019efficient}, we model the posterior $q_{\phi}(\mathbf{z}|\mathbf{c}^{\mathcal{T}_i}_{1:N})$ as a product of independent Gaussian factors, where each factor is conditioned on a single transition:
\begin{equation}
q_{\phi}(\mathbf{z}|\mathbf{c}^{\mathcal{T}_i}_{1:N}) \propto \prod_{n=1}^{N}\Psi_{\phi}(\mathbf{z}|\mathbf{c}^{\mathcal{T}_i}_{n})
\end{equation}
where each factor $\Psi_{\phi}(\mathbf{z}|\mathbf{c}^{\mathcal{T}_i}_n)$ is a Gaussian distribution. The parameters of this distribution are generated by a neural network, $f_{\phi}$, which serves as the core of the encoder. Specifically, $f_{\phi}$ maps a single transition tuple $\mathbf{c}^{\mathcal{T}_i}_n$ to a mean and variance, such that the factor is defined as $\mathcal{N}(f_{\phi}^{\mu}(\mathbf{c}^{\mathcal{T}_i}_n), f_{\phi}^{\sigma}(\mathbf{c}^{\mathcal{T}_i}_n))$. This formulation results in a final posterior that is also a tractable Gaussian distribution, whose parameters are computed by aggregating the parameters of the individual factors.

\subsubsection{Policy optimization}
We train the actor and critic networks with the on-policy PPO algorithm. The agent first collects a batch of transitions by executing its policy $\pi_{\theta}(\mathbf{a}|\mathbf{o}, \mathbf{z})$, which is conditioned on the current observation $\mathbf{o}$ and the latent variable $\mathbf{z}$ inferred from the corresponding task-specific context buffer. These transitions are stored in a temporary on-policy rollout buffer $\mathcal{B}$. After the rollout, we compute advantages and returns with these collected transitions and update the policy. 

The actor network $\pi_{\theta}(\mathbf{a}|\mathbf{o}, \mathbf{z})$ is updated by maximizing the PPO clipped surrogate objective. The critic network $Q_{\psi}(\mathbf{o}, \mathbf{a}, \mathbf{z})$ is updated by minimizing the mean squared error (MSE) against the computed returns. Upon completion of the update phase, the on-policy buffer $\mathcal{B}$ is cleared.

The design and training objective for the encoder $q_{\phi}(\mathbf{z}|\mathbf{c})$ are detailed in Section \ref{Predictive Context Encoder}.

\subsection{Predictive Context Encoder} \label{Predictive Context Encoder}
The preceding sections \ref{prob state} and \ref{Meta-RL} introduce the conceptual role of the context encoder $q_{\phi}(\mathbf{z}|\mathbf{c})$. This section presents our novel implementation of this component, termed the predictive context encoder. This encoder is specifically designed to infer a $D$-dimensional task-specific variable $\mathbf{z}$ by optimizing a multi-objective loss function.

Distinct from traditional value-aware methods trained on an implicit critic signal \cite{rakelly2019efficient}, our predictive context encoder learns from direct supervisory signals to explicitly predict one-step system dynamics and immediate rewards. This direct supervision fosters a more structured latent representation, improving both sample efficiency and training stability.

The total loss for the encoder $\mathcal{L}_{\text{encoder}}$ is a weighted sum of a regularizing KL divergence loss $\mathcal{L}_{\text{KL}}$, a prediction loss $\mathcal{L}_{\text{pred}}$, and a composite specialization loss $\mathcal{L}_{\text{spec}}$:
\begin{equation}
    \mathcal{L}_{\text{encoder}} =  \omega_{\text{KL}}' \mathcal{L}_{\text{KL}} + \mathcal{L}_{\text{pred}} + \omega_{\text{spec}} \mathcal{L}_{\text{spec}}
\end{equation}
where $\mathcal{L}_{\text{KL}}$ acts as an information bottleneck, regularizing the inferred posterior $q_{\phi}(\mathbf{z}|\mathbf{c})$ towards a prior $p(\mathbf{z}) = \mathcal{N}(0, I)$ to constrain the mutual information between the context and the latent variable, thus mitigating overfitting to the specific contexts seen during training:
\begin{equation}
    \mathcal{L}_{\text{KL}} = \mathbb{E}_\mathbf{c} \left[D_{KL}(q_{\phi}(\mathbf{z}|\mathbf{c}) \ || \ p(\mathbf{z})) \right]
\end{equation}
and the weight of this KL divergence loss $\omega_{\text{KL}}'$ is adapted dynamically during training to enhance stability.

The prediction loss $\mathcal{L}_{\text{pred}}$ is designed to make the latent variable $\mathbf{z}$ informative by ensuring it captures task-relevant dynamic information. It forces $\mathbf{z}$ to be predictive of future outcomes, comprising two distinct components:
\begin{equation}
    \mathcal{L}_{\text{pred}} = \omega_{\text{pos}}\mathcal{L}_{\text{pos}} + \omega_{\text{rew}}\mathcal{L}_{\text{rew}}
\end{equation}

The position loss $\mathcal{L}_{\text{pos}}$ and the reward loss $\mathcal{L}_{\text{rew}}$ are supervised by dedicated prediction heads. The dynamics head $f_{dyn}$ models the predicted state difference $\Delta \hat{\mathbf{o}}$ by conditioning on the current observation, action, and latent variable:
\begin{equation}
    \Delta \hat{\mathbf{o}} = f_{dyn}(\mathbf{o}, \mathbf{a}, \mathbf{z})
\end{equation}
where the function $f_{dyn}$ is implemented as a multi-layer perceptron (MLP) with fixed parameters, serving purely as a complex non-linear mapping whose weights are not updated during training. This prediction is then trained by minimizing an MSE against the ground truth difference of the positional component, denoted by the subscript $(\cdot)_{\text{pos}}$, as position is the primary state variable affected by the system's dynamic variations and is critical for successful navigation:
\begin{equation}
    \mathcal{L}_{\text{pos}} = \mathbb{E} \left[ \left\| (\Delta \hat{\mathbf{o}})_{\text{pos}} - (\mathbf{o}' - \mathbf{o})_{\text{pos}} \right\|^2 \right]
\end{equation}

Similarly, the reward head $f_{rew}$, an MLP that takes the observation and latent variable as input, computes the predicted reward $\hat{r}$:
\begin{equation}
    \hat{r} = f_{rew}(\mathbf{o}, \mathbf{z})
\end{equation}
This prediction is supervised by a robust Huber loss to mitigate the effect of outliers:
\begin{equation}
    \mathcal{L}_{\text{rew}} = \mathbb{E} \left[ \text{Huber}(\hat{r}, r) \right]
\end{equation}

The specialization loss $\mathcal{L}_{\text{spec}}$ promotes latent variable diversity across tasks, to prevent representation collapse, where the encoder might learn to ignore $\mathbf{z}$ or map all tasks to a single point. The loss is defined as:
\begin{equation}
    \mathcal{L}_{\text{spec}} = \text{clamp}(-\log(\text{Var}(\mathbf{z}) + \epsilon), L_{\min}, L_{\max})
\end{equation}

\subsection{Policy Training and Deployment} \label{sec:policy training}
The overall framework is illustrated in Fig. \ref{training pipeline}. The policy is trained using the PPO algorithm \cite{PPO}. The policy network is an MLP with two 128-unit hidden layers, followed by a linear output layer with a $\tanh$ activation to produce normalized actions. To better satisfy the Markov property, the input observation is augmented with the action from the previous timestep. The context encoder is a smaller MLP with two 64-unit hidden layers. Its final layer outputs the parameters of the Gaussian posterior distribution by aggregating information from the sampled context batch. The context buffer for each task stores up to 256 transitions. The encoder is updated periodically every $N_{\text{enc}}$ steps via gradient accumulation. During each update, gradients are computed and backpropagated for each task individually on a sampled batch of $N$ transitions from the buffer, then accumulated before a single optimization step is performed. The training procedure is detailed in Algorithm \ref{alg:meta-training}.

\begingroup
\setlength{\textfloatsep}{1em}
\begin{algorithm}[]
\caption{Meta-Training Procedure}
\label{alg:meta-training}
\begin{algorithmic}[1]
\State \textbf{Initialize:} Actor $\pi_{\theta}$, Critic $Q_{\psi}$, Encoder $q_{\phi}$.
\State \textbf{Initialize:} On-policy rollout buffer $\mathcal{B}$, off-policy context buffers $\{\mathcal{C}_i\}$.

\For{each meta-training iteration}
    \For{\textit{step} $n$ in training steps}
        \For{each task $\mathcal{T}_i$}
            \State Sample context batch $\mathbf{c}^{\mathcal{T}_i}_{1:N} \sim \mathcal{C}_i$.
            \State Infer latent variable $\mathbf{z}_i \sim q_{\phi}(\mathbf{z}_i|\mathbf{c}^{\mathcal{T}_i}_{1:N})$.
            \State Execute policy $\mathbf{a}_n \sim \pi_{\theta}(\mathbf{a}_n|\mathbf{o}_n, \mathbf{z}_i)$.
            \State Obtain reward $r_n$ and next observation $\mathbf{o}'_n$ .
            \State Add context $\mathbf{c}_{n}^{\mathcal{T}_i} = (\mathbf{o}_n,\mathbf{a}_n,r_n,\mathbf{o}'_n)$ to $\mathcal{B}$. 
            \State Add a random subset of $\mathbf{c}_{n}^{\mathcal{T}_i}$ to $\mathcal{C}_i$.
            
        \EndFor

        \If{$n \bmod N_{\text{enc}} = 0$}
            \State Compute accumulated $\mathcal{L}_{\text{encoder}}$.
            \State Update Encoder $q_{\phi}$ using $\mathcal{L}_{\text{encoder}}$.
        \EndIf
    \EndFor
    \State Compute advantages and returns for all data in $\mathcal{B}$.
        \For{each mini-batch from $\mathcal{B}$}
            \State Update Actor $\pi_{\theta}$ and Critic $Q_{\psi}$.
        \EndFor
    \State Clear on-policy storage: $\mathcal{B} \leftarrow \emptyset$.
\EndFor
\end{algorithmic}
\end{algorithm}
\endgroup

Besides, all observations are normalized prior to network processing: the quadrotor relative positions $\Delta\mathbf{p} \gets \Delta\mathbf{p} / \mathbf{k}_p$, the quadrotor velocity $\mathbf{v} \gets \mathbf{v} / \mathbf{k}_v$, where the division is conducted element-wise, and $\mathbf{k}_p$, $\mathbf{k}_v$ are normalization constants listed in Table~\ref{tab:Simulation_Parameters}. 

\begin{table}[b]
    % \vspace{-1.0em}
    \caption{Simulation and Training Parameters}
    % \vspace{-0.5em}
    \begin{center}
        \renewcommand{\arraystretch}{1.2}
        \begin{tabular}{ll|ll}
            \Xhline{1pt}
            \textbf{Parameter} & \textbf{Value} & \textbf{Parameter} & \textbf{Value} \\ \hline
            $m_{min},m_{max} $ [kg] & $0.25, 0.50$ & $\delta_{min},\delta_{max}$ [\%] & $0, 50$  \\
            $W$ & $2$ & $\boldsymbol{\omega}_{\text{max}}$ [rad/s] & $[12, 12, 5]$ \\
            $\mathbf{k}_p$ [m] & $[6, 6, 1]$ & $\mathbf{k}_v$ [m/s] & $[15, 15, 10]$ \\ 
            $\lambda_1$ & $10$ & $\lambda_2$ & $1 \times 10^{-4}$ \\ 
            $\lambda_3$ & $10$ & $r_{\text{collision}}$ & $1$ \\ 
            $\omega_{\text{pos}}$ & $1.0$ & $\omega_{\text{rew}}$ & $1.0$\\
            $\omega_{\text{spec}}$ & $5 \times 10^{-3}$ & $\alpha_{\text{scale}}$ & $0.05$ \\
            $\beta_{\text{scale}}$ & $0.1$ & $D$ & $6$\\
            $N$ & $128$ &$N_{\text{enc}}$ & $3$ \\
            \Xhline{1pt}
        \end{tabular}
    \end{center}
    \label{tab:Simulation_Parameters}
\end{table}

Meta-RL typically requires extensive training time—often hours or even days—to learn across a distribution of tasks. To minimize this, we leverage Genesis \cite{Genesis}, a state-of-the-art, GPU-vectorized simulator that supports a massive number of parallel environments for physics-based quadrotor simulation. To further reduce the required training time, we train all tasks concurrently. This is achieved by assigning each task to its own distinct set of 4096 parallel environments, allowing the agent to gather diverse experiences from all tasks simultaneously in each training step. Notably, the task-specific context buffer is populated with transitions collected from a random subset of its dedicated environments at each timestep. For the mass variation scenario, we create 16 tasks by sampling the quadrotor mass uniformly from a range of 0.25 kg to 0.5 kg. For the actuator thrust loss scenario, we create 20 tasks by simulating a thrust loss in one of the four rotors. The magnitudes of thrust loss are set to 10\%, 20\%, 30\%, 40\% and 50\%. All training is performed on a workstation equipped with an AMD Ryzen R9-9950 CPU and an NVIDIA RTX 5090 D GPU. This setup allows the policies for these two scenarios to converge after approximately 4.92 and 7.37 billion timesteps, which corresponds to only 35 and 53 minutes of training time, respectively. 

The training is conducted with a 0.01s timestep, with waypoints sampled from a [-3.0, 3.0] $\times$ [-3.0, 3.0] $\times$ [0.5, 1.5] m workspace. The policy receives only the next two waypoints as input. A waypoint is considered passed when the quadrotor enters a 1.0m acceptance radius, which triggers the waypoint update. Specifically, the future waypoint becomes the new immediate target, and a new future waypoint is randomly sampled. This process is repeated until a terminal condition is met, such as a collision or reaching the maximum episode length. Consequently, this method is not confined to several fixed tracks but facilitates continuous navigation along arbitrary multi-point tasks by dynamically updating waypoints. Correspondingly, the policy's relative position observations, $\Delta \mathbf{p}_1$ and $\Delta \mathbf{p}_2$, are computed based on these two updated waypoints.

To ensure high fidelity and seamless alignment with the physical platform, the control architecture is designed to replicate the real-world setup: the policy network outputs collective throttle and angular rate commands. These commands are then processed by a simulated Betaflight autopilot, matching the experimental hardware used in Section \ref{section: experiment setup and result}. This module incorporates a standard PID loop followed by a motor mixer, translating the policy's outputs into individual motor RPM commands, which ultimately drive the quadrotor dynamics.

During deployment, the trained encoder and policy networks are frozen and utilized directly, with no gradient updates. As the agent interacts with the environment, it maintains an online context buffer $\mathcal{C}$ of recent experiences. At each decision step, the encoder infers the latent variable $\mathbf{z}$ from this buffer. This inference is the sole mechanism for adaptation: as the buffer accumulates more transitions, the estimate $\mathbf{z}$ more accurately captures the current environment's dynamics. The policy then conditions on this refined $\mathbf{z}$ and the current observation $\mathbf{o}$ to generate agile flight commands, enabling real-time online adaptation to unknown quadrotor masses or thrust losses. The procedure is detailed in Algorithm \ref{alg:meta-testing}.

\begingroup
\setlength{\textfloatsep}{1em}
\begin{algorithm}[]
\caption{Meta-Testing Procedure (Deployment)}
\label{alg:meta-testing}
\begin{algorithmic}[1]
\State \textbf{Given:} Trained actor $\pi_{\theta}$, trained encoder $q_{\phi}$.
\State \textbf{Initialize:} Empty online context buffer $\mathcal{C} \leftarrow \emptyset$.
\State \textbf{Initialize:} Prior distribution $p(\mathbf{z}) = \mathcal{N}(0, I)$.
\State Get initial observation $\mathbf{o}_0$.

\For{each step $t$ in the episode}
    % \State \Comment{1. Infer task identity from all available context}
    \If{$\mathcal{C}$ is empty}
        \State Sample from prior: $\mathbf{z} \sim p(\mathbf{z})$.
    \Else
        \State Infer from context: $\mathbf{z} \sim q_{\phi}(\mathbf{z}|\mathcal{C})$.
    \EndIf
    
    % \State \Comment{2. Act using the inferred task identity}
    \State Execute policy $\mathbf{a}_t \sim \pi_{\theta}(\mathbf{a}_t|\mathbf{o}_t, \mathbf{z})$.
    
    % \State \Comment{3. Update context with new transition}
    \State Obtain reward $r_t$ and next observation $\mathbf{o}_{t+1}$.
    \State Add transition $(\mathbf{o}_t, \mathbf{a}_t, r_t, \mathbf{o}_{t+1})$ to online buffer $\mathcal{C}$.
\EndFor
\end{algorithmic}
% \vspace{-1em}
\end{algorithm}
% \vspace{-1.5em}
\endgroup

\section{Simulation Results and Analysis}
\begin{table*}[b]
  \centering
  \caption{Comparison of Our Method against Baselines Across Various Drone Masses. Metrics include average velocity (Avg Vel.), flight time (Time), and the percentage of flight duration where the normalized throttle exceeds 0.8 (Thr. $>$ 0.8). Results for the mass-specific RL policy are shown in bold. Crashed trials are denoted by a dash (-).}
  \label{tab:sim_line}  
  \small
  \setlength{\tabcolsep}{2.5pt}
  \sisetup{
    table-format=2.2,
    detect-weight,
    mode=text
  }
  \begin{tabular}{
  l
  @{\hspace{6pt}}
  S S[table-format=2.1] S 
  @{\hspace{16pt}}
  S S[table-format=2.1] S 
  @{\hspace{16pt}}
  S S[table-format=2.1] S 
  @{\hspace{16pt}}
  S S[table-format=2.1] S
}
    \toprule
    \multirow{3}{*}{\textbf{Method~}} &

    \multicolumn{3}{c@{\hspace{16pt}}}{\textbf{260g}} &
    \multicolumn{3}{c@{\hspace{16pt}}}{\textbf{330g}} &
    \multicolumn{3}{c@{\hspace{16pt}}}{\textbf{440g}} &
    \multicolumn{3}{c}{\textbf{550g}} \\ 
    \cmidrule(l{0em}r{1.5em}){2-4} \cmidrule(l{0em}r{1.5em}){5-7} \cmidrule(l{0em}r{1.5em}){8-10} \cmidrule(l{0em}r{0em}){11-13}
    & {Avg Vel.} & {Thr $>$ 0.8} & {Time} & {Avg Vel.} & {Thr. $>$ 0.8} & {Time} & {Avg Vel.} & {Thr. $>$ 0.8} & {Time} & {Avg Vel.} & {Thr. $>$ 0.8} & {Time} \\
    & {(m/s)} & {(\%)} & {(s)} & {(m/s)} & {(\%)} & {(s)} & {(m/s)} & {(\%)} & {(s)} & {(m/s)} & {(\%)} & {(s)} \\
    \midrule
    RL-260g~        & \bfseries 7.82 & \bfseries 71.0 & \bfseries 1.54 &  7.07 & 82.1 &  1.67 & {-}  & {-}  & {-} & {-}  & {-}  & {-} \\
    RL-330g~        & 5.47 & 61.0 & 2.12 & \bfseries 6.04 & \bfseries 77.4 & \bfseries 1.67 & 5.96 & 86.9 & 1.90 & {-}  & {-}  & {-}  \\
    RL-440g~        & 6.37 & 62.2 & 1.97 & 6.65 & 81.0 & 1.67 & \bfseries 6.06 & \bfseries 94.2 & \bfseries 1.78 & {-}  & {-}  & {-} \\
    RL-550g~        & 4.93 & 35.0 & 3.16 & 5.57 & 59.2 & 2.10 & 5.34 & 82.1 & 2.06 & \bfseries 5.23 & \bfseries 96.9 & \bfseries 2.12 \\
    RL-DR~             & 6.18 & 47.3 & 2.63 & 6.16 & 73.6 & 2.53 & 5.86 & 87.1 & 1.85 & 5.04 & 99.4 & 3.07 \\
    \specialrule{0.05em}{0.5ex}{0.5ex}
    \textbf{OURS~} & 7.31 & 69.3 & 1.56 & 6.53 & 80.5 & 1.67 & 5.93 & 90.0 & 1.79 & 5.42 & 99.1 & 2.13 \\
    \bottomrule
  \end{tabular}
\end{table*}

\begin{figure*}[b]
    \centering
% \vspace{-0.5em}
    \includegraphics[width=1\textwidth, trim=0 350 0 0, clip]{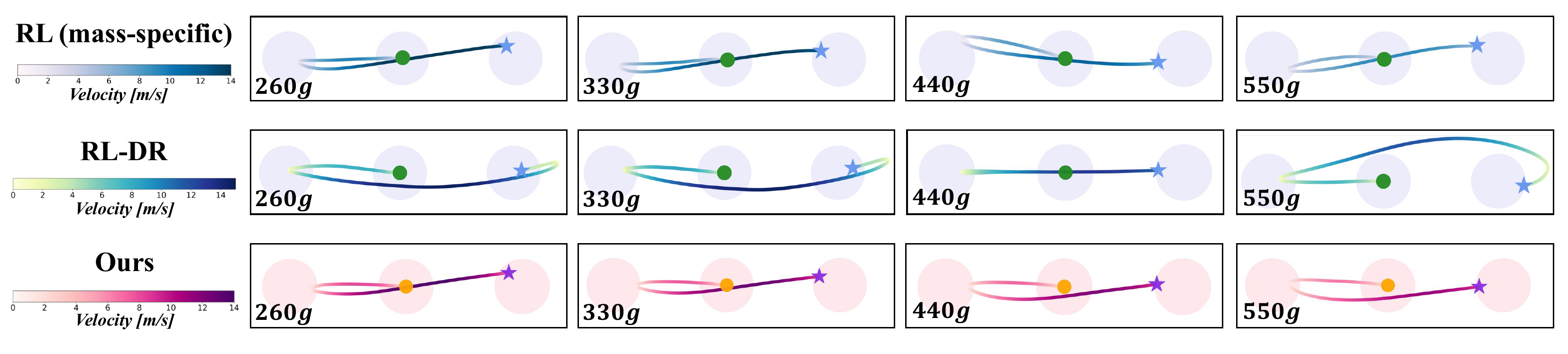}
    \caption{Comparison of trajectories on a switchback track for quadrotors with varying mass (260g, 330g, 440g, and 550g). Shaded areas denote waypoints with a radius of 1.0m. Spots and stars denote the starting and ending waypoints, respectively. Although both RL-DR and our method use a single policy, our approach yields trajectories similar to the mass-specific RL, whereas RL-DR exhibits unnecessary detours at certain masses.}
    \label{sim_line}
\end{figure*}

This section presents a comprehensive simulation study to validate the performance of our meta-RL framework. In the mass variation scenario, we evaluate the policy's effectiveness and robustness by deploying it on four quadrotors with different total masses to perform waypoint traversal on multiple tracks. In the actuator thrust loss scenario, we assess its performance against five levels of thrust loss over 100 randomly generated tracks, yielding a statistical validation of its success rate and flight efficiency. Both scenarios include tests on out-of-distribution dynamics to evaluate the policy's generalization capabilities.

\subsection{Simulation setups and baselines}
All experimental evaluations focus on the agile waypoint traversal. For each track, the quadrotor is tasked with navigating through a sequence of sparse waypoints. Utilizing the navigation policy, the quadrotor generates step-by-step motion commands based on this waypoint information, ultimately realizing a continuous, agile trajectory. To ensure consistency with the learning phase, the distance between consecutive waypoints is maintained within the spatial bounds of the training workspace. The primary objective is to traverse these waypoints as rapidly as possible, minimizing the total flight time.

In all simulation experiments, our method is benchmarked against the following two baselines: Standard RL and RL-DR. To ensure a fair comparison, all methods utilize the identical policy network architecture, the same reward function, and are trained under the same conditions (e.g., total training steps, parallel environments, etc.):

\begin{itemize}
\item \textbf{Standard RL:} A standard PPO policy trained only on the nominal quadrotor configuration. This baseline represents the optimal performance achievable in the nominal environment. It serves as a crucial benchmark to quantify the severe performance degradation when a non-adaptive policy encounters significant dynamic variations.
\item \textbf{RL-DR:} A policy trained with domain randomization (DR) across the entire distribution of variations, similar to the approach in \cite{ferede2025one}. This baseline represents a prominent RL method for robust, high-speed quadrotor flight. The cited work, for instance, trains a single neural network using DR to perform high-speed drone racing across physically distinct platforms. Critically, this baseline and the cited work address the complete end-to-end navigation and control problem, not just low-level control.
\end{itemize}

\subsection{Mass variation}

\begin{figure*}[htp]
    \centering
    \includegraphics[width=0.97\linewidth, trim=30 50 150 0]{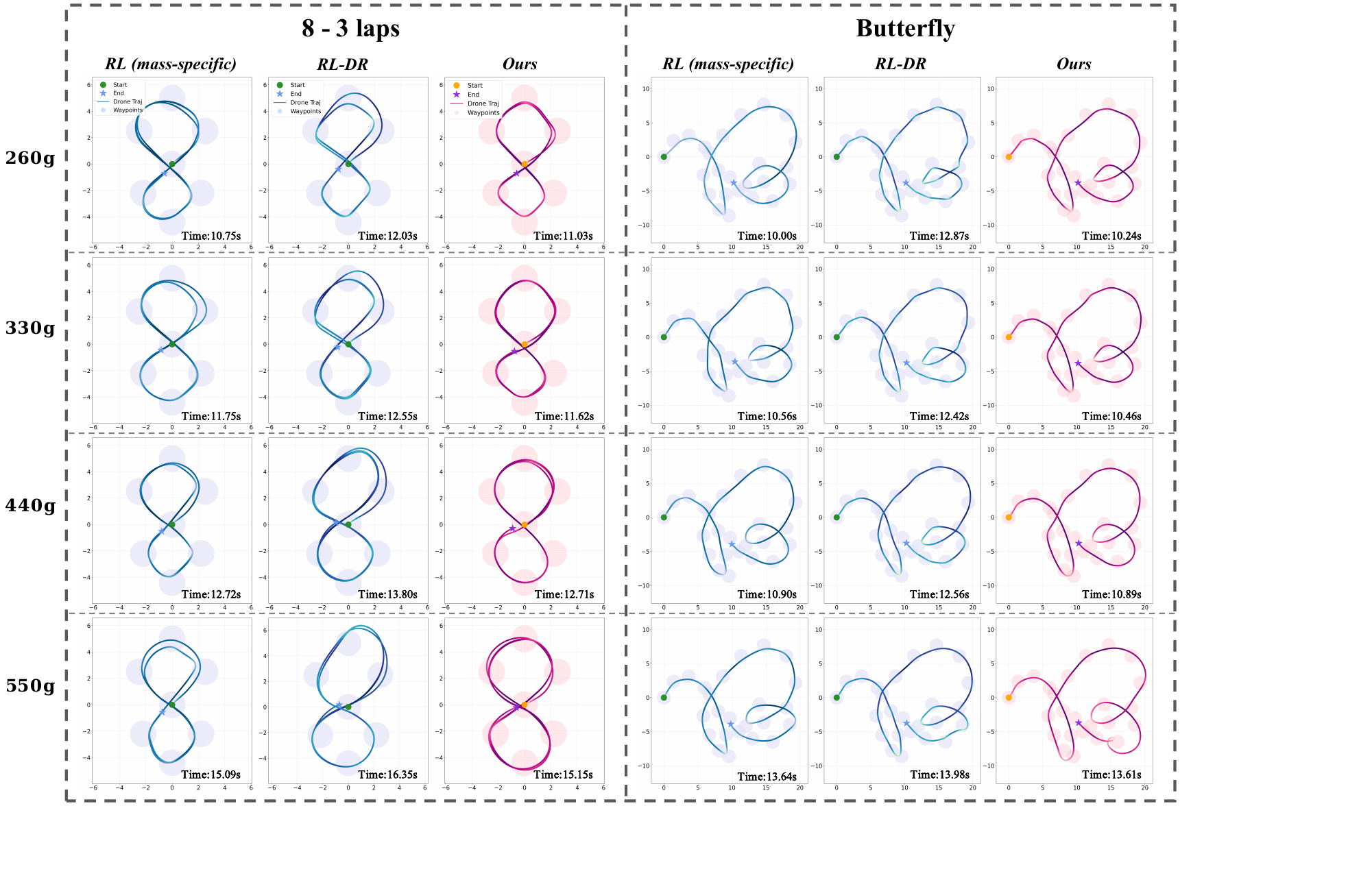}
    % \vspace{-0.5em}
    \caption{Flight trajectories and completion times for our policy and two baselines (mass-specific RL and RL-DR) on two challenging tracks. Our method's performance closely resembles the mass-specific expert's, avoiding the inefficient detours or braking of the RL-DR.}
    \label{mass-tracks}
\end{figure*}

We validate the adaptability of our meta-RL framework to varying quadrotor masses by evaluating its agile flight performance on several distinct tracks. Each track is tested on quadrotors with four distinct masses: 260g, 330g, 440g, and 550g. All tests share the same maximum thrust, which corresponds to a nominal maximum thrust-to-weight ratio of 3.5 for the 330g quadrotor. Notably, the 550g mass is outside the training distribution. We compare our method against the previously defined baselines. The standard RL baseline is implemented as a collection of four distinct 'expert' policies, each trained exclusively for one of the four tested masses (260g, 330g, 440g, and 550g). The RL-DR baseline is a single policy trained with domain randomization, covering a mass range from 0.25kg to 0.50kg.

% \textcolor{blue}{constant twr}

The initial evaluation is conducted on a challenging switchback track. In this track, the quadrotor is required to fly from a starting waypoint at (0,0,1) m, pass through an intermediate waypoint at (-3,-3,1) m, and finally reach the ending waypoint at (3,3,1) m, necessitating aggressive deceleration and rapid re-acceleration to execute the acute-angle turn. A visual comparison of the flight trajectories is provided in Fig. \ref{sim_line}, and detailed quantitative results are summarized in Table \ref{tab:sim_line}.

These results reveal a clear distinction in adaptation strategies. An expert RL policy achieves high efficiency on its specific mass but fails to generalize, exhibiting a sharp performance degradation on alternate masses due to significant model mismatch, often resulting in severe position overshoot or catastrophic failure. Our meta-RL framework, using a single policy, rapidly infers and adapts to the specific mass to achieve near-expert performance. Conversely, the RL-DR policy learns a single, conservative strategy robust to all masses, thereby sacrificing peak performance. This suboptimality is particularly evident with lighter quadrotors, where the RL-DR policy applies high throttle less often than our policy, constrained by the need to maintain stability for heavier configurations and thus failing to fully exploit the quadrotor's dynamic potential.

To further validate these findings on long-horizon missions requiring sustained agility, we present the resulting trajectories and completion times of more demanding, larger-scale tracks in Fig. \ref{mass-tracks}. These tracks include a complex '8 - 3laps' track consisting of 24 waypoints and a 'butterfly' track consisting of 23 waypoints. Although both our method and RL-DR use a single policy to navigate quadrotors of varying masses, our approach demonstrates superior efficiency, closely matching the mass-specific RL with higher speeds and shorter flight times, confirming that our adaptability advantages scale effectively to complex, multi-waypoint navigation tasks.

\subsection{Thrust loss}
\begin{figure*}[h]
    \centering
    \subfigure[Top-down view of flight trajectories with velocity profiles.]
    {\includegraphics[width=0.48\textwidth, trim=0 30 265 0, clip]{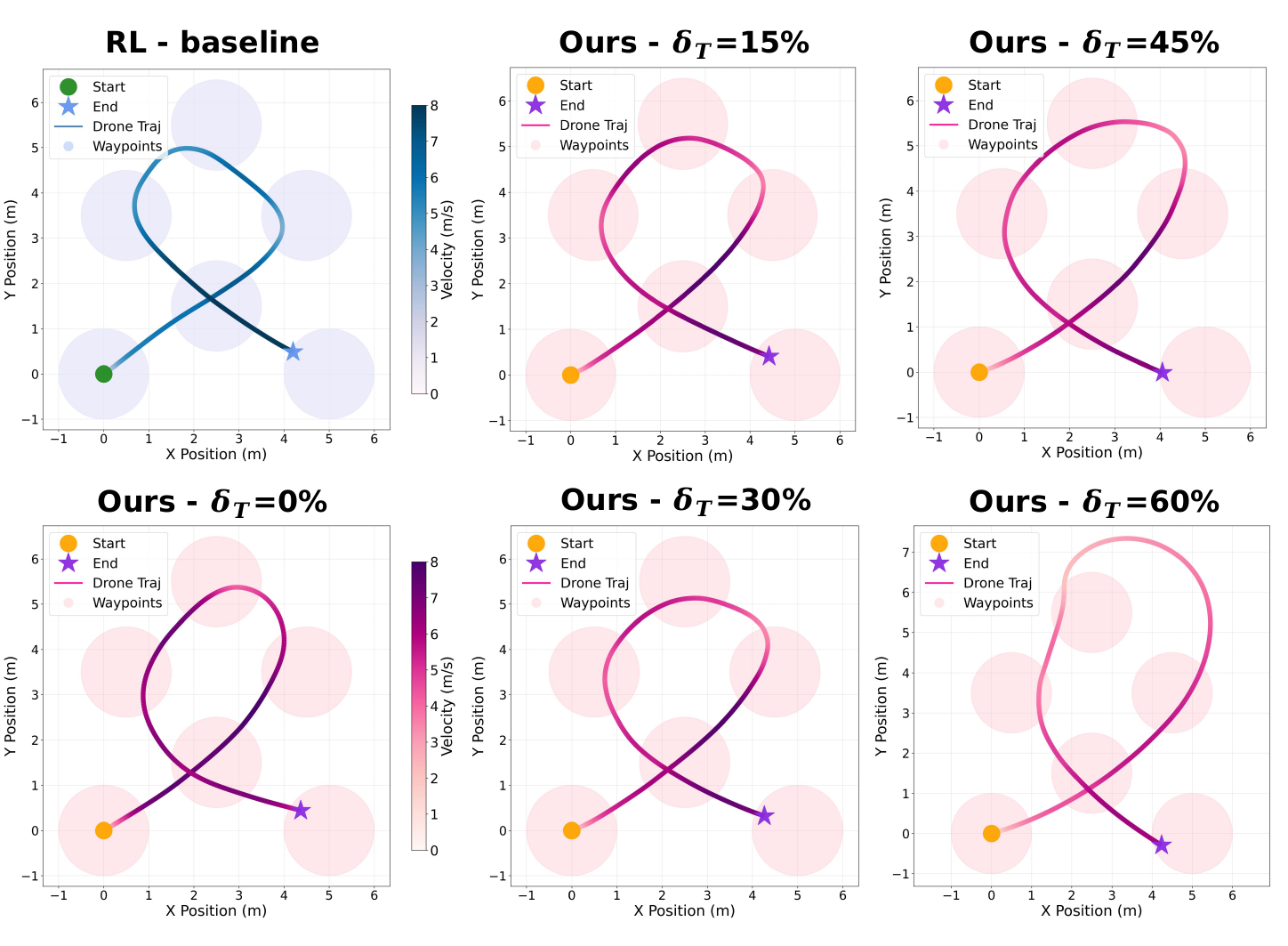}}
    \subfigure[Time series of the quadrotor position (x,y).]{\includegraphics[width=0.51\textwidth, trim = 0 0 0 0, clip]{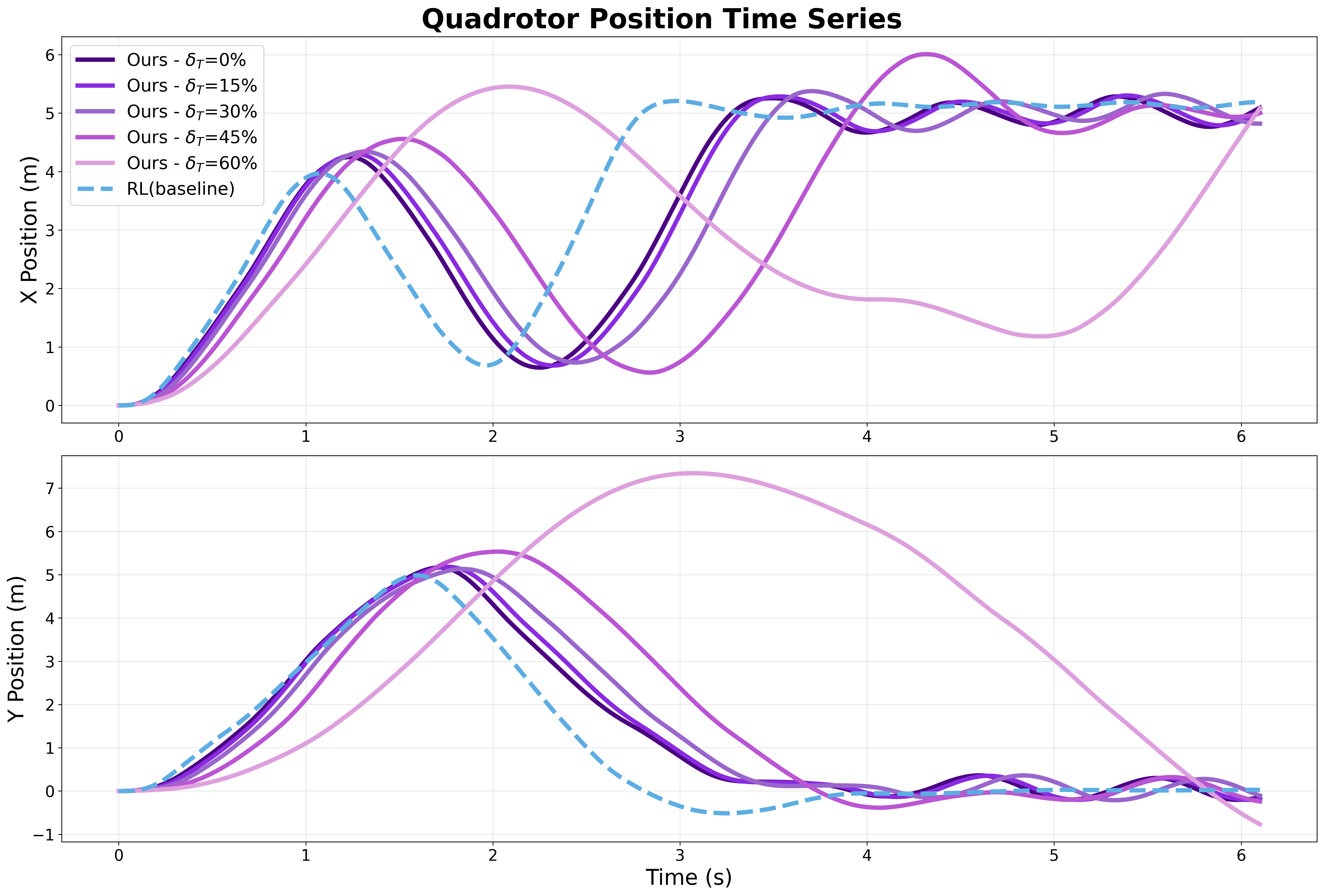}}
    \caption{Trajectories of our method against varying degrees of thrust loss, with the result of a standard RL method as a baseline.}
    \label{thrust series}
\end{figure*}
\begin{figure}[h]
    \centering
    \subfigure[]
    {\includegraphics[width=0.46\textwidth, trim=0 0 0 0, clip]{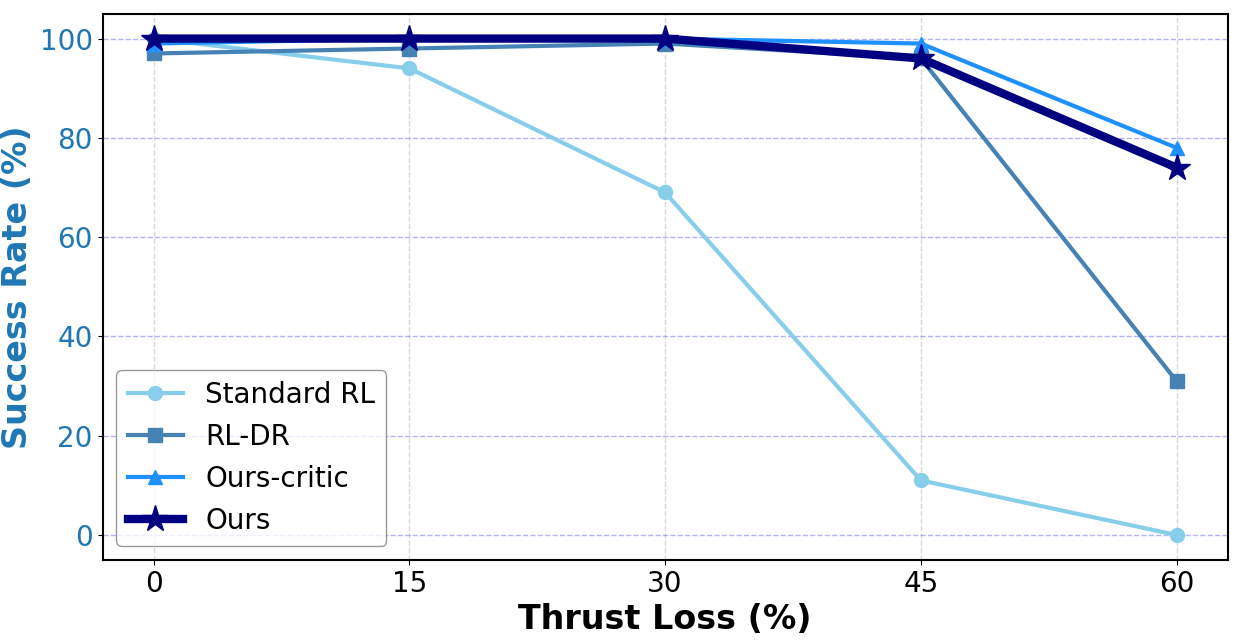}}
    \subfigure[]
    {\includegraphics[width=0.46\textwidth, trim = 0 0 0 0, clip]{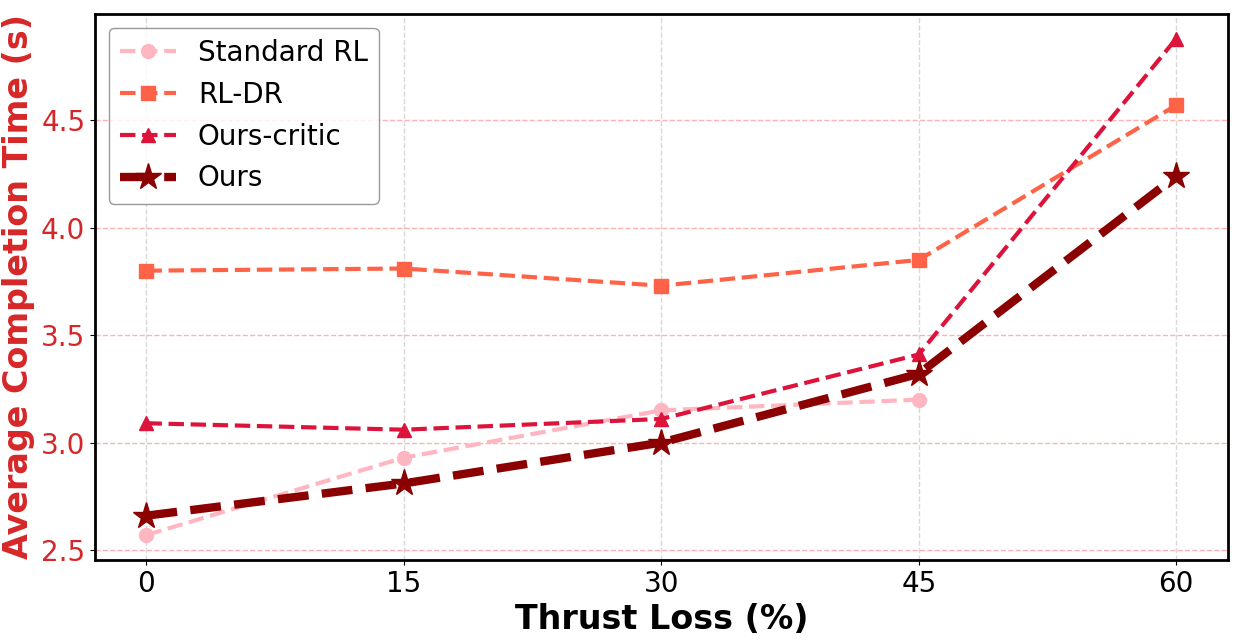}}
    \caption{Performance comparison of the proposed method against baselines under varying single-rotor thrust loss. (a) indicates the success rate (\%), while (b) shows the average completion time (s).}
    \label{curve chart}
\end{figure}

We designed an extensive simulation study to validate our framework's fault tolerance, specifically its ability to perform zero-shot online adaptation to severe, unforeseen actuator faults. The study is conducted on a 330g quadrotor with a maximum thrust-to-weight ratio of 3.5.

We illustrate our method's performance against varying degrees of thrust loss on a representative track. Since all waypoints are set to a constant altitude of 1.0m, we present the X-Y planar trajectory along with the time-series position plots for each axis, as shown in Fig. \ref{thrust series}. Notably, our method successfully navigates a quadrotor suffering up to 60\% thrust loss on a single rotor, which lies outside the training range of [0\%, 50\%] while maintaining a high degree of agility across all tested cases. This visualization confirms that the policy is not just passively surviving, but actively compensating for the fault to maintain high performance.

We further assess our policy and the baseline methods across 100 randomly generated 5-waypoint tracks inside the same workspace used during training. For this specific fault-tolerance scenario, the standard RL baseline was trained only on the nominal quadrotor, specifically the 0\% thrust loss case, while the RL-DR baseline was trained using domain randomization across the entire fault distribution, spanning a range from 0\% to 50\% thrust loss. Additionally, to benchmark the performance of our specific encoder design, we compare it against a critic-based encoder from \cite{rakelly2019efficient} on the same set of tracks. Performance is quantified by success rates and average completion time under five distinct single-rotor thrust loss scenarios, namely 0\%, 15\%, 30\%, 45\%, and 60\%, where the 60\% case provides an out-of-distribution test of generalization. The results are presented in Fig. \ref{curve chart}.

The simulation results demonstrate our method's superior fault tolerance capability against severe, unobserved thrust loss. The standard RL policy proves brittle, as expected, since its fixed strategy cannot compensate for altered dynamics. Its success rate collapses to 69\% at a 30\% thrust loss and then drops further to 11\% at 45\% loss, and ultimately fails completely. While the RL-DR policy improves robustness, maintaining a high success rate above 96\% for thrust loss up to 45\%, the rate plummets to just 31\% at the out-of-distribution loss level. This highlights the limitation of a purely robust policy, which is 'averaged' across the training distribution and fails when the fault is too extreme.

In stark contrast, our meta-RL policies demonstrate a superior combination of performance and robustness, maintaining a near-perfect success rate with completion time closer to the expert's than RL-DR, and still succeeding in over 70\% of trials under a severe 60\% loss. This success demonstrates the power of online adaptation: the policy actively infers the fault from recent experience and adjusts its behavior accordingly. Our ablation study compares our predictive context encoder against the critic-based variant. While both methods prove effective, the results highlight a difference in efficiency. Our predictive encoder consistently yields shorter completion times across all fault conditions, outperforming the critic-based variant.

\section{Experiment Setup and Result}\label{section: experiment setup and result}
Consistent with the simulation study, we conduct real-world experiments to validate our method in both mass variation and actuator thrust loss scenarios. These experiments are also designed to evaluate the sim-to-real transfer capability of our trained policies.
\subsection{Experiment Setup}
\begin{figure}[h!]
    \centering
    \includegraphics[width=0.44\textwidth, trim=0 50 370 0, clip]{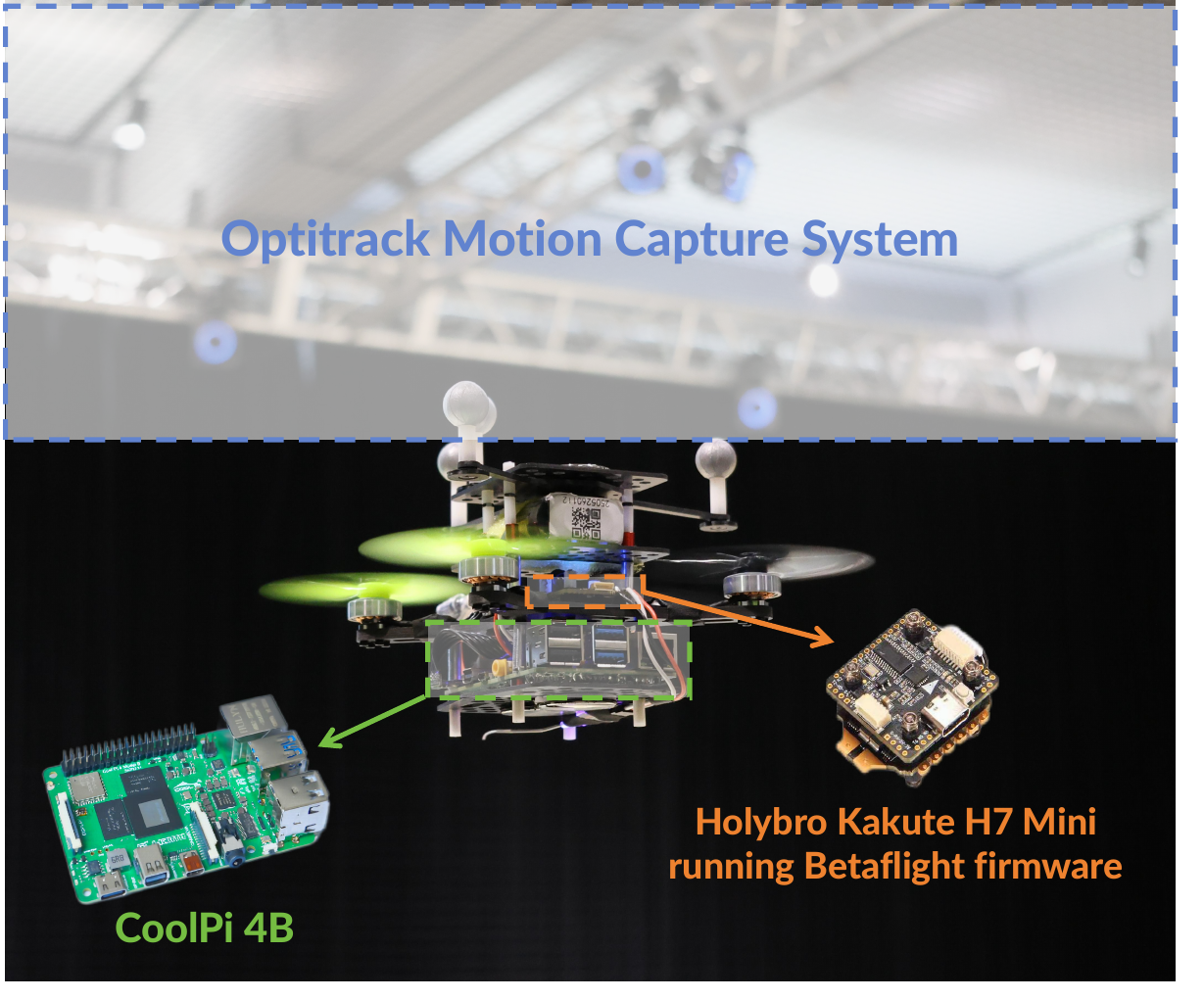}
    \caption{Experimental platform.}
    \label{fig:platform}
\end{figure}

Our experimental platform, shown in Fig. \ref{fig:platform}, is a custom-built quadrotor with a nominal mass of 330g and a maximum thrust-to-weight ratio of 3.5 without additional payload. This nominal configuration serves as the basis for all subsequent dynamic variations.

The quadrotor's onboard electronics consist of two primary components: a 10g autopilot running the Betaflight open-source firmware and a Cool Pi onboard computer. The autopilot serves as the low-level controller, managing both the PID-based angular rate loop and the motor mixer. The Cool Pi is responsible for high-level computation, running both the policy network and the context encoder.

\begin{figure}
    \centering
    \includegraphics[width=0.4\textwidth, trim=0 260 550 0, clip]{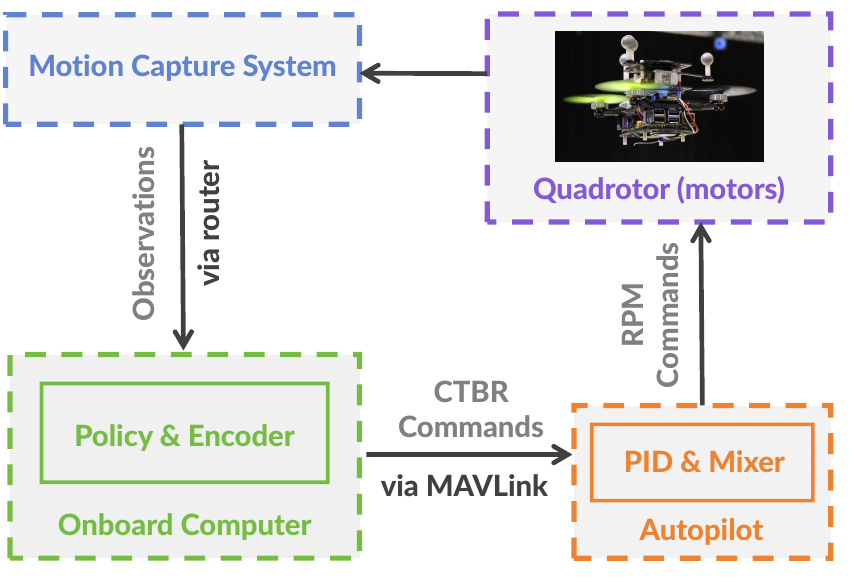}
    \caption{Data flow in the experiment.}
    \label{fig:signal}
\end{figure}

For data flow (Fig. \ref{fig:signal}), we use a motion capture system to provide accurate position and attitude measurements to the Cool Pi. The policy network, deployed using LibTorch, runs at 100Hz to predict desired rate and throttle commands. These commands are then sent to the autopilot via the MAVLink protocol for motor control. To ensure consistency, as mentioned in Section \ref{sec:policy training}, our simulation environment implements an identical mixer logic to translate policy outputs (throttle and rate) into motor RPM commands, mirroring the functionality of the Betaflight firmware. Notably, simulation results confirm that this low-level module, by itself, cannot adequately compensate for large dynamic variations.

\subsection{Experiment Results}
\subsubsection{Mass variation}
In the mass variation scenario, we demonstrate the policy's true online adaptation capability by utilizing a single policy to perform three consecutive flights along the same 7-waypoint track without landing or policy switching. This setup is designed to be a challenging, real-time test of the system's inference mechanism.

Specifically, after the quadrotor completes a flight and enters a stable hover, we alter its total mass by attaching magnet payloads, as shown in Fig. \ref{real_mass_var}. The total masses of the quadrotor for three flights are 330g, 440g, and 550g, respectively. This procedure forces the quadrotor to recognize and adapt to unobserved changes in its own dynamics purely by processing its recent flight experience through the context encoder.

\begin{figure}[h]
    \centering
    \includegraphics[width=0.49\textwidth, trim=0 23 220 0, clip]{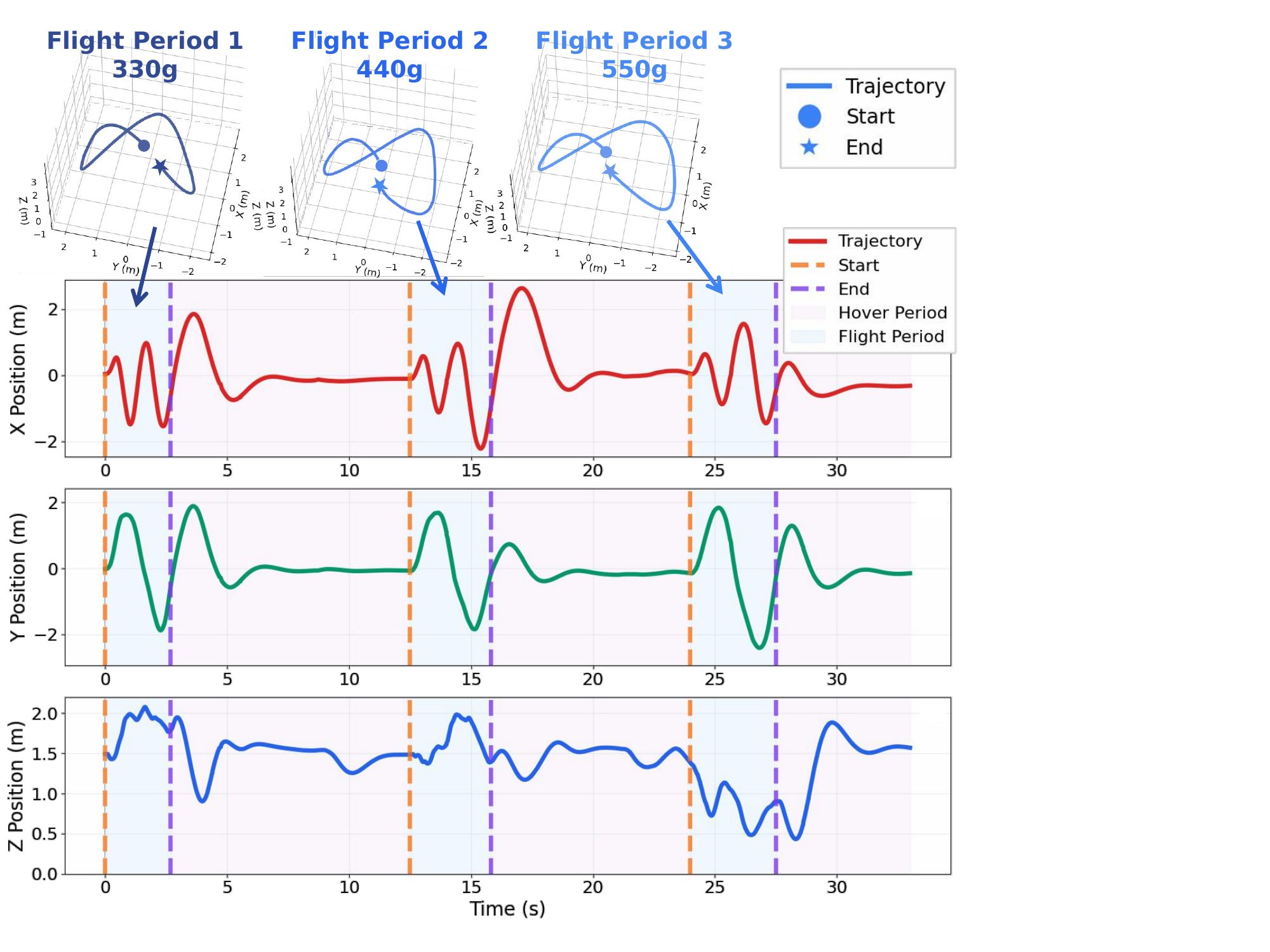}
    % \vspace{-0.5em}
    \caption{ Real-world 3D trajectories and time-series plots of three consecutive flights without landing in the mass variation scenario.}
    \label{real mass}
\end{figure}

The learned policy demonstrates excellent adaptability to these significant variations. As shown in Fig. \ref{real mass}, the three consecutive flight trajectories remain remarkably consistent. This consistency is the key finding: it visually confirms that the policy successfully inferred the new dynamics online and immediately compensated by adjusting its control strategy to maintain the desired path.

Furthermore, the correspondence between our simulation and real-world experiments is quantitatively confirmed in Table \ref{tab:flight_metrics_transposed}. This not only underscores the method's zero-shot sim-to-real transfer capability but also proves its ability to maintain high agility despite a severe mass increase of up to 66.7\%.

\begin{table}[htbp]
\centering
\caption{Comparison of Simulation and Real-world Flight Metrics Across Mass Variations.}
\label{tab:flight_metrics_transposed}

\begin{tabular}{l *{6}{S[table-format=1.2]}}
\toprule
\multirow{2}{*}{\textbf{Performance Metric}} & \multicolumn{2}{c}{\textbf{330g}} & \multicolumn{2}{c}{\textbf{440g}} & \multicolumn{2}{c}{\textbf{550g}} \\
\cmidrule(lr){2-3} \cmidrule(lr){4-5} \cmidrule(lr){6-7}
& {\textbf{Sim}} & {\textbf{Real}} & {\textbf{Sim}} & {\textbf{Real}} & {\textbf{Sim}} & {\textbf{Real}} \\
\midrule
Completion Time (s) & 2.50 & 2.68 & 2.98 & 3.22 & 3.47 & 3.52 \\
Max Velocity (m/s)  & 7.29 & 7.69 & 6.77 & 7.11 & 6.07 & 6.48 \\
\bottomrule
% \vspace{-1.5em}
\end{tabular}
\end{table}

\subsubsection{Thrust loss}
\begin{figure*}[h]
    \centering
    % \vspace{-0.3em}
    \includegraphics[width=0.82\textwidth, trim=0 80 130 0, clip]{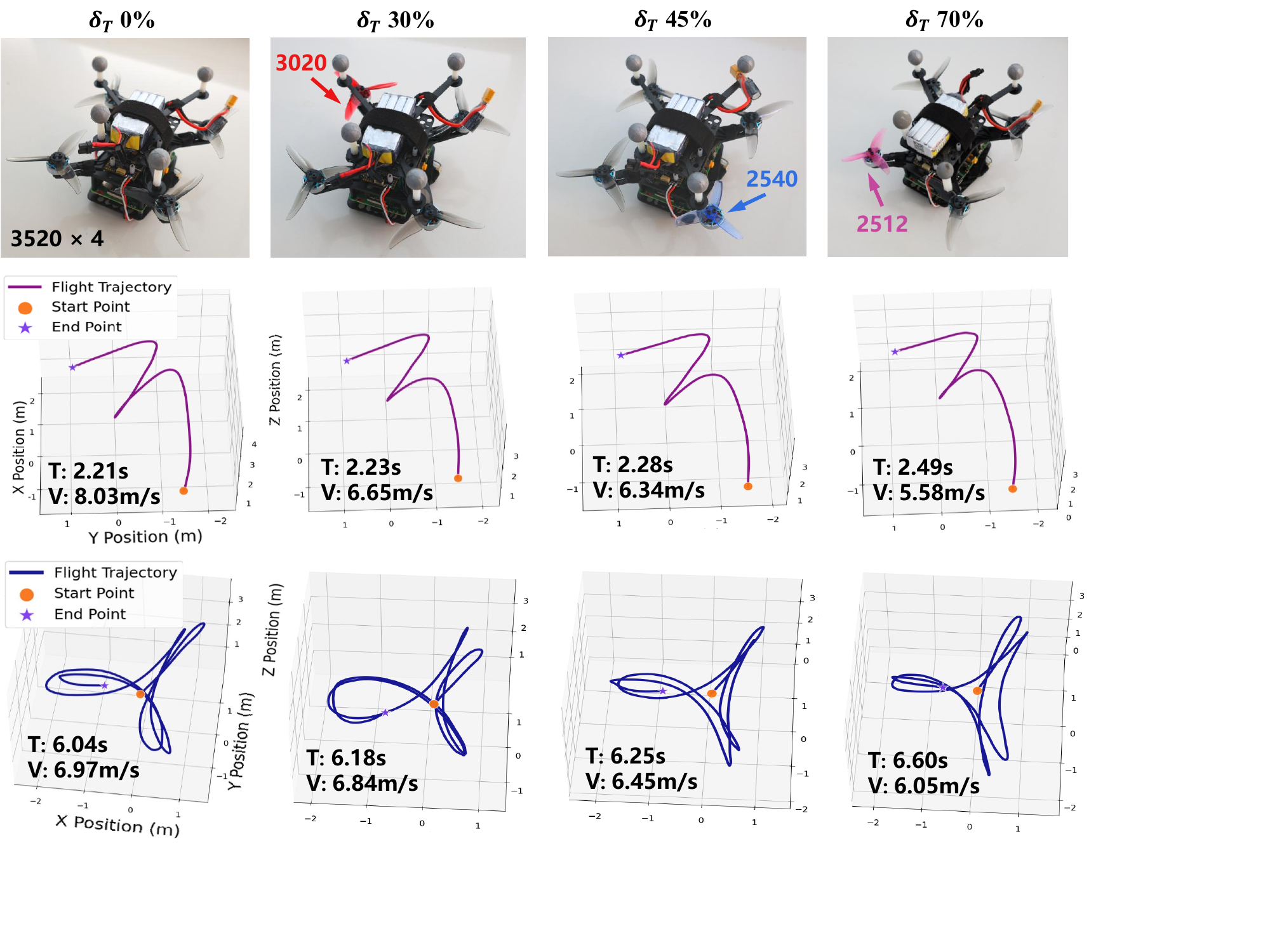}
    \caption{ Results of the thrust loss scenario. Upper: our real-world setup, where a four-digit label specifies a propeller's diameter and pitch in tenths of an inch. Bottom: flight trajectories for the 'M' and 'A' tracks, annotated with total flight time (T) and maximum velocity (V).}
    \label{real thrust}
    % \vspace{-1.2em}
\end{figure*}

For the actuator thrust loss scenario, we conducted a challenging real-world study to validate the policy's zero-shot online adaptation to a severe, unforeseen fault. This was achieved by intentionally replacing a standard propeller with a smaller one to create a significant, unobserved thrust loss, as shown in Fig. \ref{real thrust}.

We selected three propellers that, at a typical hovering thrust, induce thrust losses of approximately 30\%, 45\%, and 70\%, respectively. Crucially, the 70\% loss level is severe and lies outside the training distribution, which only goes up to 50\%. To validate the policy's robustness to the fault's location, not just its magnitude, the fault-inducing propeller was mounted at a different rotor position for each set of trials. This ensures the policy must actively infer the specific dynamic anomaly, such as a roll or yaw bias, rather than relying on an overfit assumption of which motor failed.

For each fault condition, we execute two representative flights: a 5-waypoint 'M'-shaped track and a 13-waypoint 'A'-shaped track. The resulting trajectories are shown in Fig. \ref{real thrust}. Notably, the policy maintains high-speed, safe flight across all tested fault severities. Even when facing the severe 70\% thrust loss, a condition far too significant to be mitigated by the low-level autopilot alone, our policy successfully completes the tracks, exhibiting only a minor degradation in agility. This success highlights the effectiveness of the online inference mechanism, which enables the policy to actively compensate for the fault in real-time.
\section{CONCLUSIONS}
In this work, we present MAVEN, a meta-RL framework that enables a single policy to master agile flight across quadrotors with varying dynamics. By leveraging a predictive context encoder, our method achieves rapid online adaptation to unobserved variations in mass and actuator thrust loss. Our experimental results confirm that MAVEN achieves robust adaptation while retaining superior peak performance over baseline methods. The successful zero-shot sim-to-real transfer demonstrates the framework's readiness for real-world deployment. Furthermore, the proposed framework is extensible, supporting future research on adaptation to other dynamic variations like external environmental factors or various platforms.

% \bibliography{reference}

\begin{thebibliography}{10}
\providecommand{\url}[1]{#1}
\csname url@samestyle\endcsname
\providecommand{\newblock}{\relax}
\providecommand{\bibinfo}[2]{#2}
\providecommand{\BIBentrySTDinterwordspacing}{\spaceskip=0pt\relax}
\providecommand{\BIBentryALTinterwordstretchfactor}{4}
\providecommand{\BIBentryALTinterwordspacing}{\spaceskip=\fontdimen2\font plus
\BIBentryALTinterwordstretchfactor\fontdimen3\font minus \fontdimen4\font\relax}
\providecommand{\BIBforeignlanguage}[2]{{%
\expandafter\ifx\csname l@#1\endcsname\relax
\typeout{** WARNING: IEEEtran.bst: No hyphenation pattern has been}%
\typeout{** loaded for the language `#1'. Using the pattern for}%
\typeout{** the default language instead.}%
\else
\language=\csname l@#1\endcsname
\fi
#2}}
\providecommand{\BIBdecl}{\relax}
\BIBdecl

\bibitem{foehn2021time}
P.~Foehn, A.~Romero, and D.~Scaramuzza, ``Time-optimal planning for quadrotor waypoint flight,'' \emph{Science robotics}, vol.~6, no.~56, p. eabh1221, 2021.

\bibitem{wang2022geometrically}
Z.~Wang, X.~Zhou, C.~Xu, and F.~Gao, ``Geometrically constrained trajectory optimization for multicopters,'' \emph{IEEE Transactions on Robotics}, vol.~38, no.~5, pp. 3259--3278, 2022.

\bibitem{song2021autonomous}
Y.~Song, M.~Steinweg, E.~Kaufmann, and D.~Scaramuzza, ``Autonomous drone racing with deep reinforcement learning,'' in \emph{2021 IEEE/RSJ International Conference on Intelligent Robots and Systems (IROS)}.\hskip 1em plus 0.5em minus 0.4em\relax IEEE, 2021, pp. 1205--1212.

\bibitem{end2024robin}
R.~Ferede, C.~De~Wagter, D.~Izzo, and G.~C. de~Croon, ``End-to-end reinforcement learning for time-optimal quadcopter flight,'' in \emph{2024 IEEE International Conference on Robotics and Automation (ICRA)}, 2024, pp. 6172--6177.

\bibitem{kaufmann2023champion}
E.~Kaufmann, L.~Bauersfeld, A.~Loquercio, M.~M{\"u}ller, V.~Koltun, and D.~Scaramuzza, ``Champion-level drone racing using deep reinforcement learning,'' \emph{Nature}, vol. 620, no. 7976, pp. 982--987, 2023.

\bibitem{song2023reaching}
Y.~Song, A.~Romero, M.~M{\"u}ller, V.~Koltun, and D.~Scaramuzza, ``Reaching the limit in autonomous racing: Optimal control versus reinforcement learning,'' \emph{Science Robotics}, vol.~8, no.~82, p. eadg1462, 2023.

\bibitem{qin2024time}
C.~Qin, J.~Chen, Y.~Lin, A.~Goudar, A.~P. Schoellig, and H.~H.-T. Liu, ``Time-optimal planning for long-range quadrotor flights: An automatic optimal synthesis approach,'' \emph{arXiv preprint arXiv:2407.17944}, 2024.

\bibitem{sun2022comparative}
S.~Sun, A.~Romero, P.~Foehn, E.~Kaufmann, and D.~Scaramuzza, ``A comparative study of nonlinear mpc and differential-flatness-based control for quadrotor agile flight,'' \emph{IEEE Transactions on Robotics}, vol.~38, no.~6, pp. 3357--3373, 2022.

\bibitem{romero2022model}
A.~Romero, S.~Sun, P.~Foehn, and D.~Scaramuzza, ``Model predictive contouring control for time-optimal quadrotor flight,'' \emph{IEEE Transactions on Robotics}, vol.~38, no.~6, pp. 3340--3356, 2022.

\bibitem{wang2025dashing}
X.~Wang, J.~Zhou, Y.~Feng, J.~Mei, J.~Chen, and S.~Li, ``Dashing for the golden snitch: Multi-drone time-optimal motion planning with multi-agent reinforcement learning,'' in \emph{2025 IEEE International Conference on Robotics and Automation (ICRA)}.\hskip 1em plus 0.5em minus 0.4em\relax IEEE, 2025, pp. 16\,692--16\,698.

\bibitem{zhang2025learning}
D.~Zhang, A.~Loquercio, J.~Tang, T.-H. Wang, J.~Malik, and M.~W. Mueller, ``A learning-based quadcopter controller with extreme adaptation,'' \emph{IEEE Transactions on Robotics}, 2025.

\bibitem{tremblay2018training}
J.~Tremblay, A.~Prakash, D.~Acuna, M.~Brophy, V.~Jampani, C.~Anil, T.~To, E.~Cameracci, S.~Boochoon, and S.~Birchfield, ``Training deep networks with synthetic data: Bridging the reality gap by domain randomization,'' in \emph{Proceedings of the IEEE conference on computer vision and pattern recognition workshops}, 2018, pp. 969--977.

\bibitem{molchanov2019sim}
A.~Molchanov, T.~Chen, W.~H{\"o}nig, J.~A. Preiss, N.~Ayanian, and G.~S. Sukhatme, ``Sim-to-(multi)-real: Transfer of low-level robust control policies to multiple quadrotors,'' in \emph{2019 IEEE/RSJ International Conference on Intelligent Robots and Systems (IROS)}.\hskip 1em plus 0.5em minus 0.4em\relax IEEE, 2019, pp. 59--66.

\bibitem{fei2020learn}
F.~Fei, Z.~Tu, D.~Xu, and X.~Deng, ``Learn-to-recover: Retrofitting uavs with reinforcement learning-assisted flight control under cyber-physical attacks,'' in \emph{2020 IEEE International Conference on Robotics and Automation (ICRA)}.\hskip 1em plus 0.5em minus 0.4em\relax IEEE, 2020, pp. 7358--7364.

\bibitem{ferede2025one}
R.~Ferede, T.~Blaha, E.~Lucassen, C.~De~Wagter, and G.~C. de~Croon, ``One net to rule them all: Domain randomization in quadcopter racing across different platforms,'' \emph{arXiv preprint arXiv:2504.21586}, 2025.

\bibitem{zhang2022learning}
D.~Zhang, A.~Loquercio, X.~Wu, A.~Kumar, J.~Malik, and M.~W. Mueller, ``Learning a single near-hover position controller for vastly different quadcopters,'' \emph{arXiv preprint arXiv:2209.09232}, 2022.

\bibitem{kim2025reinforcement}
D.~Kim, J.~D. Lee, H.~Bang, and J.~Bae, ``Reinforcement learning-based fault-tolerant control for quadrotor with online transformer adaptation,'' \emph{arXiv preprint arXiv:2505.08223}, 2025.

\bibitem{liu2024reinforcement}
X.~Liu, Z.~Yuan, Z.~Gao, and W.~Zhang, ``Reinforcement learning-based fault-tolerant control for quadrotor uavs under actuator fault,'' \emph{IEEE Transactions on Industrial Informatics}, 2024.

\bibitem{thrun1998learning}
S.~Thrun and L.~Pratt, ``Learning to learn: Introduction and overview,'' in \emph{Learning to learn}.\hskip 1em plus 0.5em minus 0.4em\relax Springer, 1998, pp. 3--17.

\bibitem{finn2017model}
C.~Finn, P.~Abbeel, and S.~Levine, ``Model-agnostic meta-learning for fast adaptation of deep networks,'' in \emph{International conference on machine learning}.\hskip 1em plus 0.5em minus 0.4em\relax PMLR, 2017, pp. 1126--1135.

\bibitem{hospedales2021meta}
T.~Hospedales, A.~Antoniou, P.~Micaelli, and A.~Storkey, ``Meta-learning in neural networks: A survey,'' \emph{IEEE transactions on pattern analysis and machine intelligence}, vol.~44, no.~9, pp. 5149--5169, 2021.

\bibitem{duan2016rl}
Y.~Duan, J.~Schulman, X.~Chen, P.~L. Bartlett, I.~Sutskever, and P.~Abbeel, ``Rl$^2$: Fast reinforcement learning via slow reinforcement learning,'' \emph{arXiv preprint arXiv:1611.02779}, 2016.

\bibitem{rakelly2019efficient}
K.~Rakelly, A.~Zhou, C.~Finn, S.~Levine, and D.~Quillen, ``Efficient off-policy meta-reinforcement learning via probabilistic context variables,'' in \emph{International conference on machine learning}.\hskip 1em plus 0.5em minus 0.4em\relax PMLR, 2019, pp. 5331--5340.

\bibitem{bhatia2023rl}
A.~Bhatia, S.~B. Nashed, and S.~Zilberstein, ``Rl$^3$: Boosting meta reinforcement learning via rl inside rl$^2$,'' \emph{arXiv preprint arXiv:2306.15909}, 2023.

\bibitem{nagabandi2018learning}
A.~Nagabandi, I.~Clavera, S.~Liu, R.~S. Fearing, P.~Abbeel, S.~Levine, and C.~Finn, ``Learning to adapt in dynamic, real-world environments through meta-reinforcement learning,'' \emph{arXiv preprint arXiv:1803.11347}, 2018.

\bibitem{mckinnon2021meta}
C.~D. McKinnon and A.~P. Schoellig, ``Meta learning with paired forward and inverse models for efficient receding horizon control,'' \emph{IEEE Robotics and Automation Letters}, vol.~6, no.~2, pp. 3240--3247, 2021.

\bibitem{chen2024meta}
C.~Chen, C.~Li, H.~Lu, Y.~Wang, and R.~Xiong, ``Meta reinforcement learning of locomotion policy for quadruped robots with motor stuck,'' \emph{IEEE Transactions on Automation Science and Engineering}, 2024.

\bibitem{wei2025meta}
M.~Wei, L.~Zheng, Y.~Wu, R.~Mei, and H.~Cheng, ``Meta-learning enhanced model predictive contouring control for agile and precise quadrotor flight,'' \emph{IEEE Transactions on Robotics}, 2025.

\bibitem{o2022neural}
M.~O’Connell, G.~Shi, X.~Shi, K.~Azizzadenesheli, A.~Anandkumar, Y.~Yue, and S.-J. Chung, ``Neural-fly enables rapid learning for agile flight in strong winds,'' \emph{Science Robotics}, vol.~7, no.~66, p. eabm6597, 2022.

\bibitem{eschmann2025raptor}
J.~Eschmann, D.~Albani, and G.~Loianno, ``Raptor: A foundation policy for quadrotor control,'' \emph{arXiv preprint arXiv:2509.11481}, 2025.

\bibitem{belkhale2021model}
S.~Belkhale, R.~Li, G.~Kahn, R.~McAllister, R.~Calandra, and S.~Levine, ``Model-based meta-reinforcement learning for flight with suspended payloads,'' \emph{IEEE Robotics and Automation Letters}, vol.~6, no.~2, pp. 1471--1478, 2021.

\bibitem{cao2024autonomous}
Q.~Cao, Z.~Liu, H.~Yu, X.~Liang, and Y.~Fang, ``Autonomous landing of the quadrotor on the mobile platform via meta reinforcement learning,'' \emph{IEEE Transactions on Automation Science and Engineering}, vol.~22, pp. 2269--2280, 2024.

\bibitem{eschmann2024learning}
J.~Eschmann, D.~Albani, and G.~Loianno, ``Learning to fly in seconds,'' \emph{IEEE Robotics and Automation Letters}, vol.~9, no.~7, pp. 6336--6343, 2024.

\bibitem{PPO}
J.~Schulman, F.~Wolski, P.~Dhariwal, A.~Radford, and O.~Klimov, ``Proximal policy optimization algorithms,'' \emph{arXiv preprint arXiv:1707.06347}, 2017.

\bibitem{Genesis}
\BIBentryALTinterwordspacing
G.~Authors, ``Genesis: A generative and universal physics engine for robotics and beyond,'' December 2024. [Online]. Available: \url{https://github.com/Genesis-Embodied-AI/Genesis}
\BIBentrySTDinterwordspacing

\end{thebibliography}
\bibliographystyle{IEEEtran}

\vfill

\end{document}